\documentclass[10pt,twocolumn,letterpaper]{article}

\usepackage{wacv}
\usepackage{times}
\usepackage{epsfig}
\usepackage{graphicx}
\usepackage{amsmath}
\usepackage{amssymb}

\usepackage{chngcntr}

\usepackage{multirow,tabularx}
\usepackage{booktabs}
\usepackage{pifont}
\usepackage{floatrow}
\usepackage{xcolor}
\usepackage{subcaption}
\usepackage{enumitem}

\newcommand{\blue}[1]{\textcolor{blue}{#1}}

\newcommand*\ttvar[1]{\texttt{\expandafter\dottvar\detokenize{#1}\relax}}
\newcommand*\dottvar[1]{\ifx\relax#1\else
  \expandafter\ifx\string_#1\string_\allowbreak\else#1\fi
  \expandafter\dottvar\fi}

\usepackage{tablefootnote}

\renewcommand\fbox{\fcolorbox{blue}{white}}

\newcommand{\datasetName}{DocVQA\xspace}
\newcommand{\squad}{SQuAD 1.1\xspace}
\newcommand{\vqa}{VQA 2.0\xspace}
\newcommand{\textvqa}{TextVQA\xspace}
\newcommand{\stvqa}{ST-VQA\xspace}
\newcommand{\lorra}{LoRRA\xspace}
\newcommand{\mc}{M4C\xspace}

\def\Plus{\texttt{+}}

\newcommand{\cmark}{\ding{51}}%
\newcommand{\xmark}{\ding{55}}%

\usepackage{hyperref}
\hypersetup{breaklinks=true,colorlinks}

\wacvfinalcopy

\begin{document}

\title{DocVQA: A Dataset for VQA on Document Images}

\author
{Minesh Mathew$^{1}$ $\qquad$  Dimosthenis Karatzas$^{2}$ $\qquad$  C.V. Jawahar$^{1}$ \\
$^1$CVIT, IIIT Hyderabad, India $\qquad$ 
$^2$Computer Vision Center, UAB, Spain$\qquad$
\\
{\tt\small minesh.mathew@research.iiit.ac.in, dimos@cvc.uab.es, jawahar@iiit.ac.in}
}

\maketitle

\begin{abstract}
We present a new dataset for Visual Question Answering (VQA) on document images called \datasetName. The dataset consists of 50,000 questions defined on 12,000+ document images. Detailed analysis of the dataset in comparison with similar datasets for VQA and reading comprehension is presented. We report several baseline results by adopting existing VQA and reading comprehension models. Although the existing models perform reasonably well on certain types of questions, there is large performance gap compared to human performance (94.36\% accuracy). The models need to improve specifically on questions where understanding structure of the document is crucial. The dataset, code and leaderboard are available at \href{docvqa.org}{docvqa.org}

\end{abstract}

\section{Introduction}

Research in Document Analysis and Recognition (DAR) is generally focused on information extraction tasks that aim to convert information in document images into machine readable form, such as character recognition~\cite{doermann2014handbook}, table extraction~\cite{kavasidis2019saliency} or key-value pair extraction~\cite{palm2017cloudscan}. Such algorithms tend to be designed as task specific blocks, blind to the end-purpose the extracted information will be used for.

Progressing independently in such information extraction processes has been quite successful, although it is not necessarily true that holistic document image understanding can be achieved through a simple constructionist approach, building upon such modules. The scale and complexity of the task introduce difficulties that require a different point of view.

In this article we introduce Document Visual Question Answering (\datasetName), as a high-level task dynamically driving DAR algorithms to conditionally interpret document images. By doing so, we seek to inspire a “purpose-driven” point of view in DAR research.
\begin{figure}[t]
\begin{center}
\begin{tabular}{p{0.95\linewidth}}
    \includegraphics[width=\linewidth,height=0.7\linewidth]{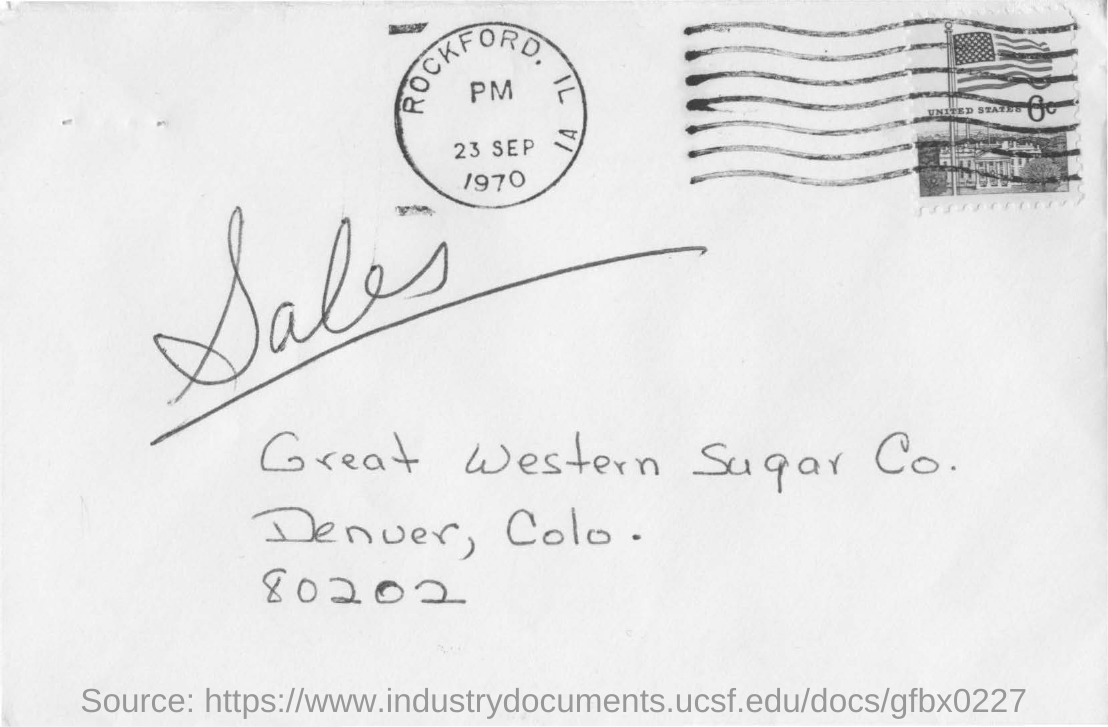}  \\

    \footnotesize{\fontfamily{qhv}\selectfont \textbf{Q:} Mention the ZIP code written? } \par \color{blue}\footnotesize{\fontfamily{qhv}\selectfont \textbf{A:} 80202} \\
    
    \footnotesize{\fontfamily{qhv}\selectfont \textbf{Q:}  What date is seen on the seal at the top of the letter? } \par 
    \color{blue}\footnotesize{\fontfamily{qhv}\selectfont \textbf{A:} 23 sep 1970} \\
    
     \footnotesize{\fontfamily{qhv}\selectfont \textbf{Q:}   Which company address is mentioned on the letter?  } \par 
    \color{blue}\footnotesize{\fontfamily{qhv}\selectfont \textbf{A:} Great western sugar Co.} \\

\end{tabular}
\end{center}
\caption{Example question-answer pairs from \datasetName. Answering questions in the new dataset require models not just to read text but interpret it within the layout/structure of the document.}
\label{fig:firspage_example}
\vspace{-5mm}

\end{figure}
In case of Document VQA, as illustrated in Figure \ref{fig:firspage_example}, an intelligent reading system is expected to respond to ad-hoc requests for information, expressed in natural language questions by human users. To do so, reading systems should not only extract and interpret the textual (handwritten, typewritten or printed) content of the document images, but exploit numerous other visual cues including layout (page structure, forms, tables), non-textual elements (marks, tick boxes, separators, diagrams) and style (font, colours, highlighting), to mention just a few. 

Departing from generic VQA \cite{vqa2} and Scene Text VQA ~\cite{textvqa,stvqa_iccv} approaches, the document images warrants a different approach to exploit all the above visual cues, making use of prior knowledge of the implicit written communication conventions used, and dealing with the high-density semantic information conveyed in such images.
Answers in case of document VQA cannot be sourced from a closed dictionary, but they are inherently open ended.

Previous approaches on bringing VQA to the documents domain have either focused on specific document elements such as data visualisations~\cite{dvqa, kahou2017figureqa} or on specific collections such as book covers \cite{mishra2019ocr}. In contrast to such approaches, we recast the problem to its generic form, and put forward a large scale, varied collection of real documents.

Main contributions of this work can be summarized as following:
\begin{itemize}[noitemsep,nolistsep]
\item We introduce \datasetName, a large scale dataset of $12,767$ document images of varied types and content, over which we have defined $50,000$ questions and answers. The questions defined are categorised based on their reasoning requirements, allowing us to analyze how \datasetName methods fare for different question types.

\item We define and evaluate various baseline methods over the \datasetName dataset, ranging from simple heuristic methods and human performance analysis that allow us to define upper performance bounds given different assumptions, to state of the art Scene Text VQA models and NLP models.
\end{itemize}

\section{Related Datasets and Tasks}

Machine reading comprehension (MRC) and open-domain question answering (QA) are two problems which are being actively pursued by Natural Language Processing (NLP) and Information Retrieval (IR) communities.
In MRC the task is to answer a natural language question given a question and a paragraph (or a single document) as the context. In case of open domain QA, no specific context is given and answer need to be found from a large collection (say Wikipedia) or from Web. 
MRC is often modelled as an extractive QA problem where answer is defined as a span of the context on which the question is defined. 
Examples of datsets for extractive QA include \squad~\cite{squad},  NewsQA~\cite{newsqa} and Natural Questions~\cite{naturalquestions}. MS MARCO~\cite{ms_marco} is an example of a QA dataset for abstractive QA where answers need to be generated not extracted. Recently Transformer based pretraining  methods like  Bidirectional Encoder Representations from Transformers (BERT)~\cite{bert} and XLNet~\cite{xlnet} have helped to build QA models outperforming Humans on reading comprehension on SQuAD~\cite{squad}. In contrast to QA in NLP where context is given as computer readable strings, contexts in case of \datasetName are document images.

Visual Question Answering (VQA) aims to provide an accurate natural language answer given an image and a natural language question. VQA has attracted an intense research effort over the past few years~\cite{vqa2, agrawal2017c, johnson2017clevr}.
Out of a large body of work on VQA, scene text VQA branch is the most related to our work.
Scene text VQA refers to VQA systems aiming to deal with cases where understanding scene text instances 
is necessary to respond to the questions posed. The \stvqa~\cite{stvqa_iccv} and \textvqa~\cite{textvqa} datasets were introduced in parallel in 2019 and were quickly followed by more research~\cite{singh2019strings, gao2020multi, wang2020general}.

The \stvqa dataset~\cite{stvqa_iccv} has $31,000\Plus$ questions over $23,000\Plus$ images collected from different public data sets. The \textvqa dataset~\cite{textvqa} has $45,000\Plus$ questions over  $28,000\Plus$ images sampled from specific categories of the OpenImages dataset~\cite{OpenImages2} that are expected to contain text.
Another dataset named OCR-VQA~\cite{mishra2019ocr} comprises more than 1 million question-answer pairs over 207K+ images of book covers. The questions in this dataset are domain specific, generated based on template questions and answers extracted from available metadata.

\begin{figure*}[ht]
    \centering
    \begin{subfigure}[t]{0.32\textwidth}
        \includegraphics[width=1\linewidth]{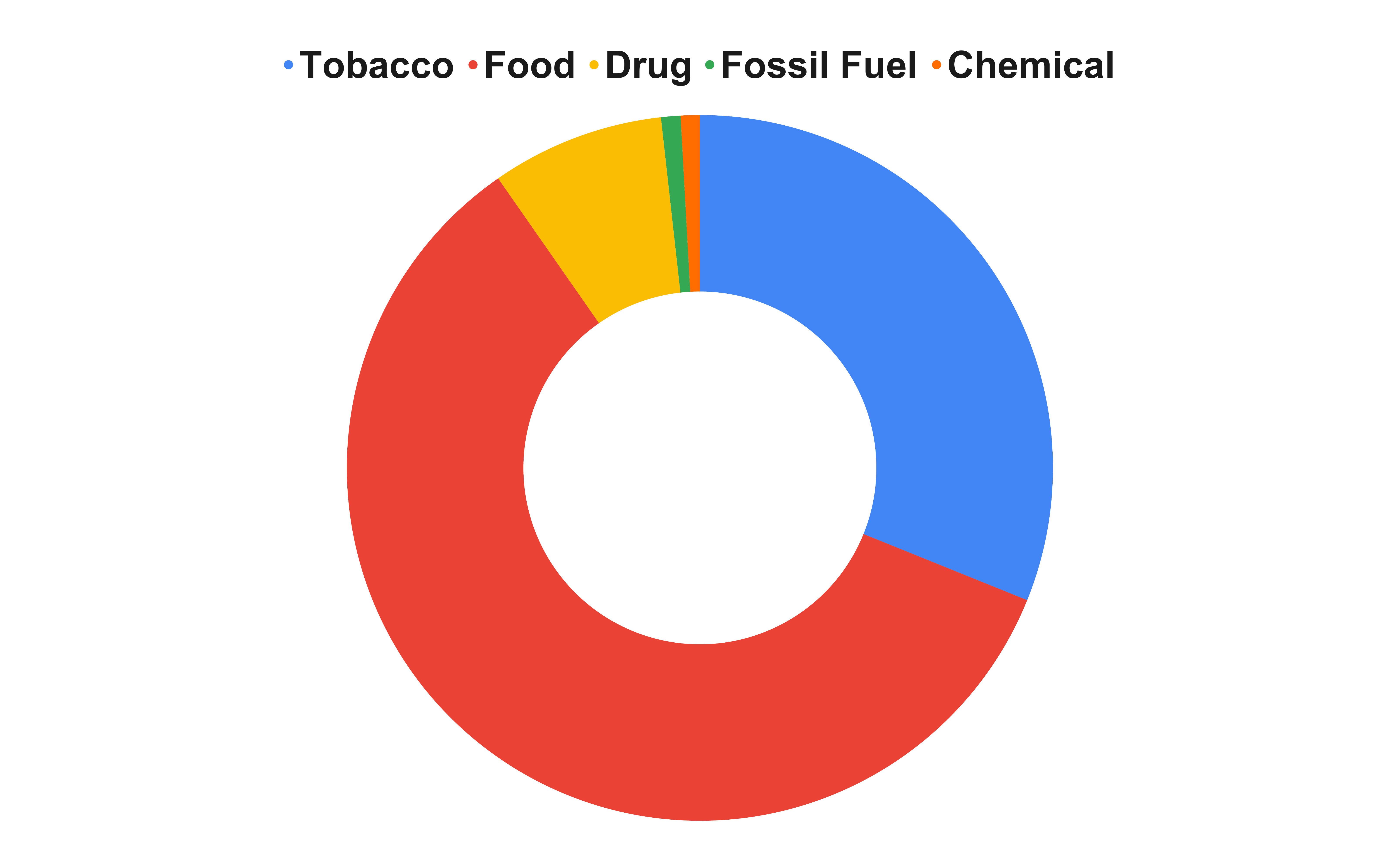}
        \caption{Industry-wise distribution of the documents.}
        \label{fig:industry_distr}
    \end{subfigure}
    \hspace{1mm}
    \begin{subfigure}[t]{0.32\textwidth}
        \includegraphics[width=1\linewidth]{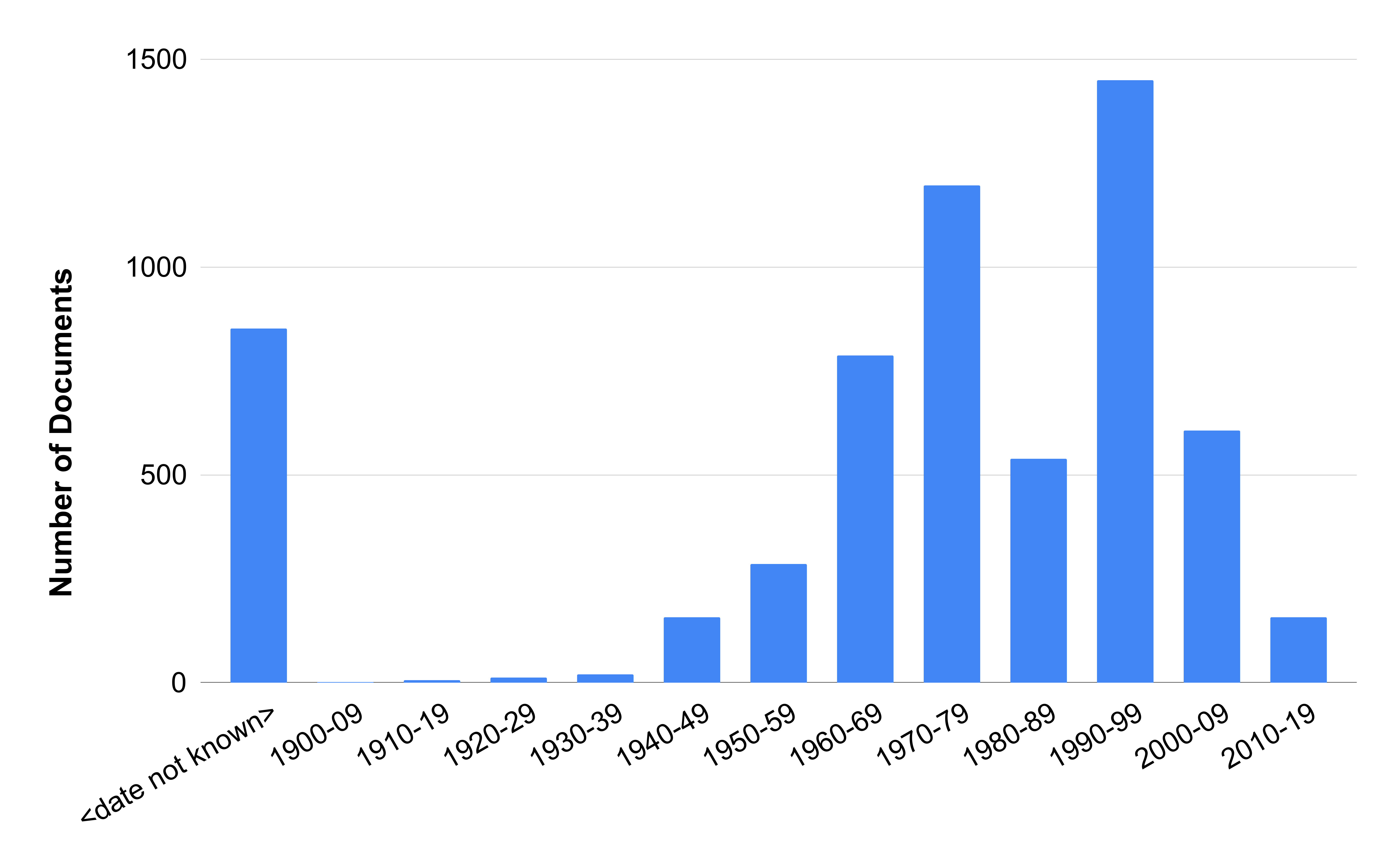}
        \caption{Year wise distribution of the documents.}
        \label{fig:doc_year_distr}
    \end{subfigure}
    \hspace{1mm}
    \begin{subfigure}[t]{0.32\textwidth}
        \includegraphics[width=1\linewidth]{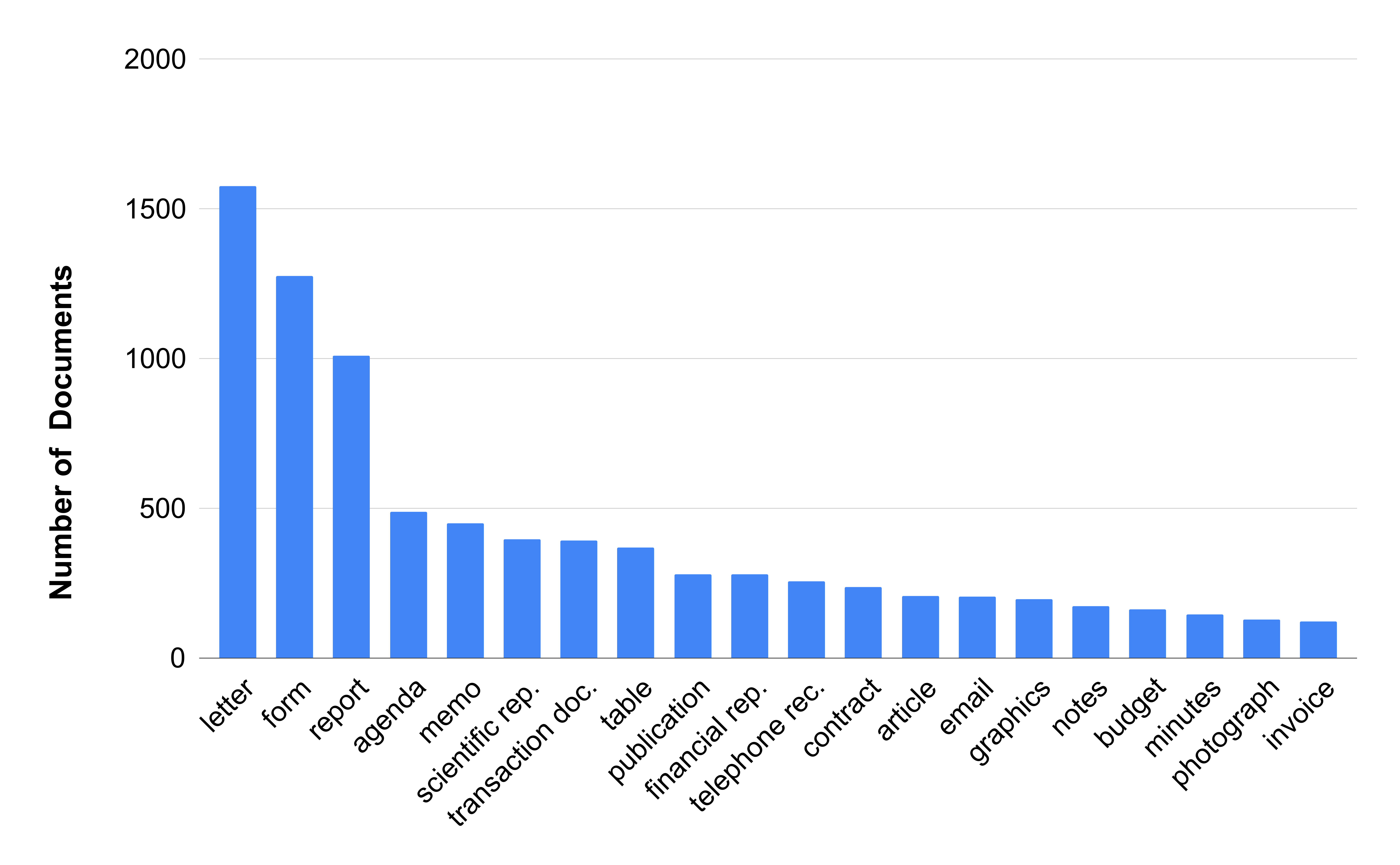}
        \caption{Various types of documents used.}
        \label{fig:doc_type_distr}
    \end{subfigure}
\caption{Document images we use in the dataset come from 6071  documents spanning many decades, of a variety of types, originating from 5 different industries. We use documents from UCSF Industry Documents Library.}
\vspace{-2mm}

\end{figure*}
Scene text VQA methods~\cite{m4c, gao2020multi, textvqa, gomez2020multimodal}
typically make use of pointer mechanisms in order to deal with out-of-vocabulary (OOV) words appearing in the image and provide the open answer space required. This goes hand in hand with the use of word embeddings capable of encoding OOV words into a pre-defined semantic space,  such as FastText~\cite{bojanowski2017enriching} or BERT~\cite{bert}. More recent, top-performing methods in this space include M4C~\cite{m4c} and MM-GNN~\cite{gao2020multi} models.

Parallelly there have been works on certain domain specific VQA tasks which require to read and understand text in the images.
The DVQA dataset presented by Kafle \etal~\cite{kafle2020answering, dvqa} comprises synthetically generated images of bar charts and template questions defined automatically based on the bar chart metadata. The dataset contains more than three million question-answer pairs over 300,000 images.

FigureQA~\cite{kahou2017figureqa} comprises over one million yes or no questions, grounded on over 100,000 images. Three different types of charts are used: bar, pie and line charts. Similar to DVQA, images are synthetically generated and questions are  generated from templates.
Another related QA task is  Textbook Question Answering (TQA)~\cite{textbookqa} where multiple choice questions are asked on multimodal context, including text, diagrams and images. Here textual information is provided in computer readable format.

Compared to these existing datasets either concerning VQA on real word images, or domain specific VQA for charts or book covers, the proposed \datasetName comprise document images.
The dataset covers a multitude of different document types that include elements like tables, forms and  figures , as well as a range of different textual, graphical and structural elements.

\section{\datasetName}
In this section we explain data collection and annotation process and present statistics and analysis of \datasetName.

\subsection{Data Collection}
\textbf{Document Images:} Images in the dataset are sourced from documents in UCSF Industry Documents Library\footnote{\url{https://www.industrydocuments.ucsf.edu/}}. 
The documents are organized under different industries and further under different collections. We downloaded documents from different collections and hand picked pages from these documents for use in the dataset. Majority of  documents in the library are binarized and the binarization has taken on a toll on the image quality. We tried to minimize binarized images in \datasetName since we did not want poor image quality to be a bottleneck for VQA.
We also prioritized pages with tables, forms, lists and figures over pages which only have running text. 

The final set of images in the dataset are drawn from pages of $6,071$ industry documents. We made use of documents from as early as 1900 to as recent as 2018. (~\autoref{fig:doc_year_distr}). Most of the documents are from the 1960-2000 period and they include typewritten, printed, handwritten and born-digital text. There are documents from all 5 major industries for which the library hosts documents --- tobacco, food, drug, fossil fuel and chemical. 
We use many documents from food and nutrition related collections, as they have a good number of non-binarized images.
. 
See ~\autoref{fig:industry_distr} for industry wise distribution of the $6071$ documents used. The documents comprise a wide variety of document types as shown in~\autoref{fig:doc_type_distr}.

\textbf{Questions and Answers:} Questions and answers on the selected document images are collected with the help of remote workers, using a Web based annotation tool. 
The annotation process was organized in three stages. In stage 1, workers were shown a document image and asked to define at most 10 question-answer pairs on it. 
We encouraged the workers to add more than one ground truth answer per question in cases where it is warranted.
\begin{figure}[b]
    \centering
    \includegraphics[width=0.8\linewidth]{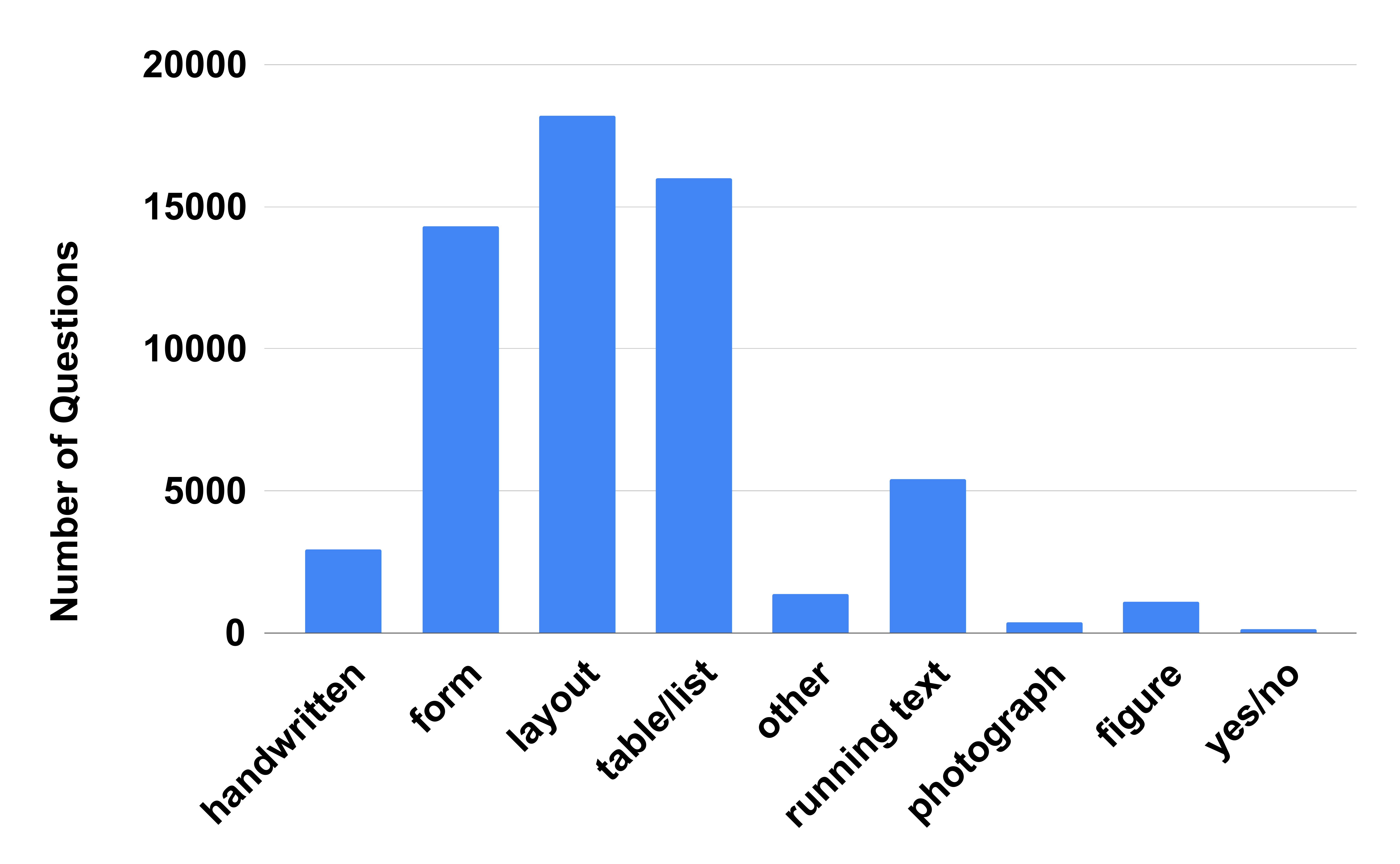}
    \caption{The 9 question types and share of questions in each type.}
    \label{fig:question_types}
\end{figure}
Workers were instructed to ask questions which can be answered using text present in the image and to enter the answer verbatim from the document. This makes VQA on the \datasetName dataset an extractive QA problem similar to extractive QA tasks in NLP~\cite{squad,newsqa} and VQA in case of \stvqa~\cite{stvqa_iccv}. 

\begin{figure*}[ht]
    \centering
    \begin{subfigure}[t]{0.32\textwidth}
        \includegraphics[width=1\linewidth]{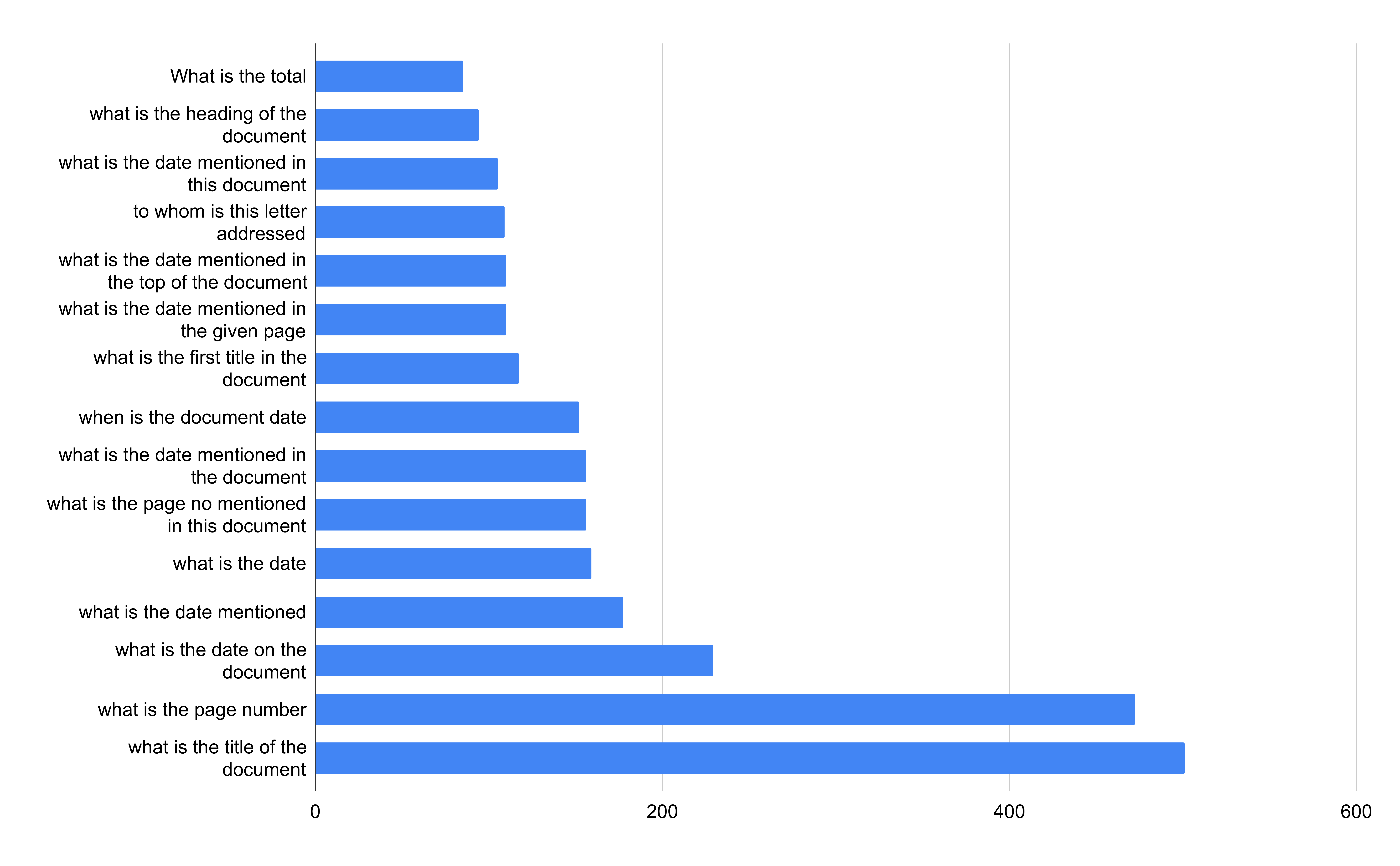}
        \caption{Top 15 most frequent questions.}
        \label{fig:top_questions}
    \end{subfigure}
    \hspace{1mm}
    \begin{subfigure}[t]{0.32\textwidth}
        \includegraphics[width=1\linewidth]{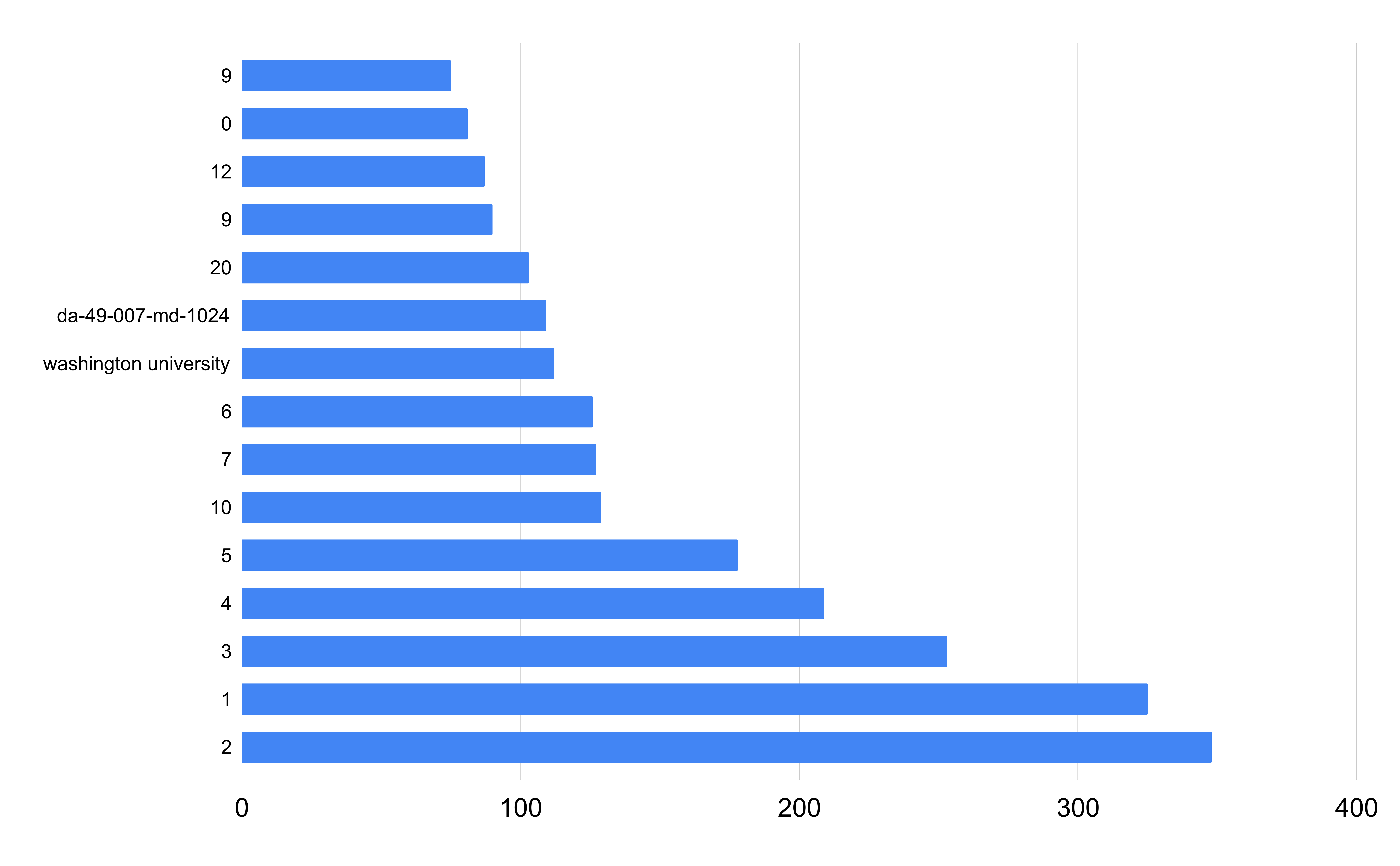}
        \caption{Top 15 most frequent answers.}
        \label{fig:top_anwers}
    \end{subfigure}
    \hspace{1mm}
    \begin{subfigure}[t]{0.32\textwidth}
        \includegraphics[width=1\linewidth]{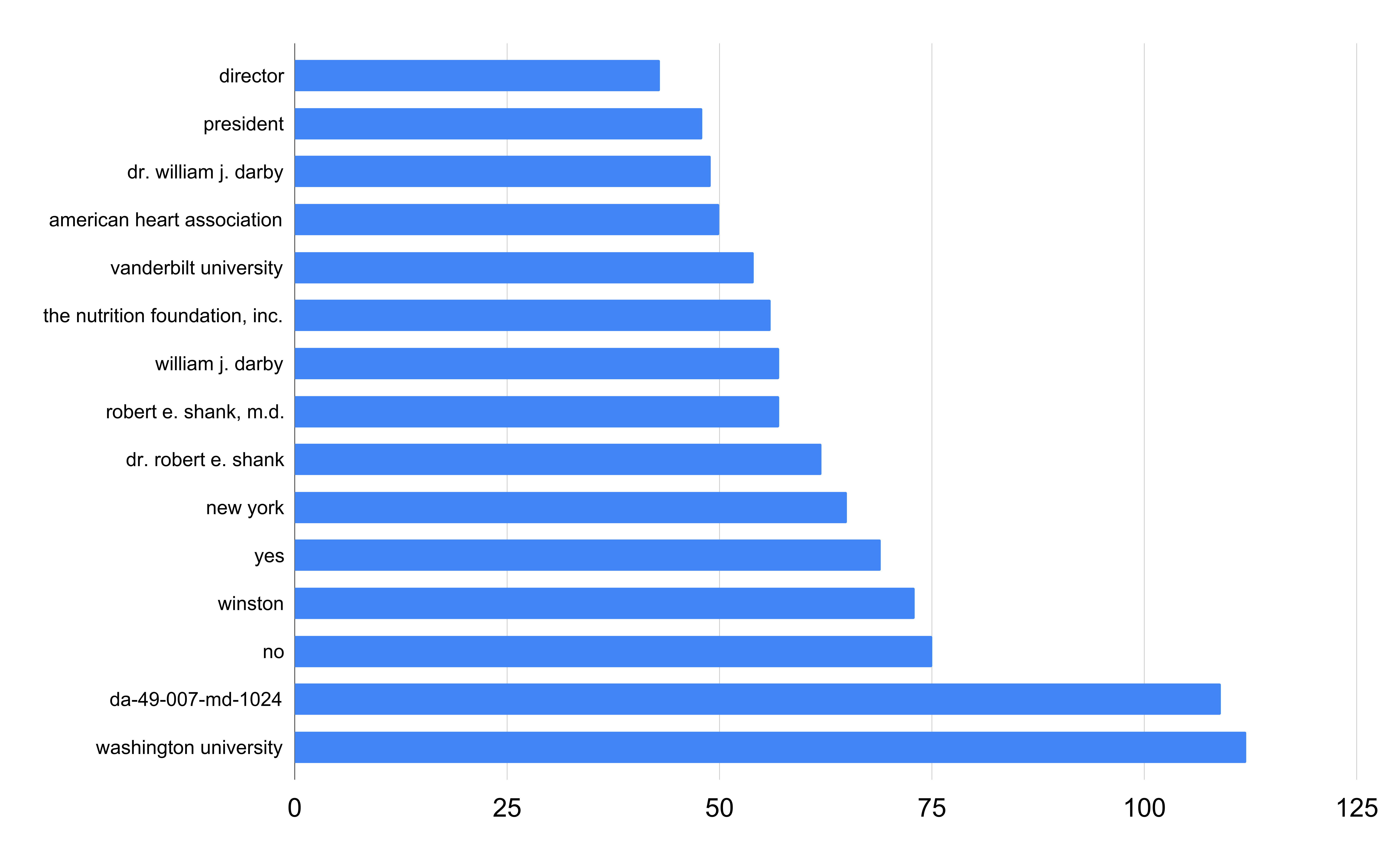}
        \caption{Top 15 non numeric answers.}
        \label{fig:top_non_numeric_Answers}
    \end{subfigure}
    \begin{subfigure}[t]{0.32\textwidth}
        \includegraphics[width=1\linewidth]{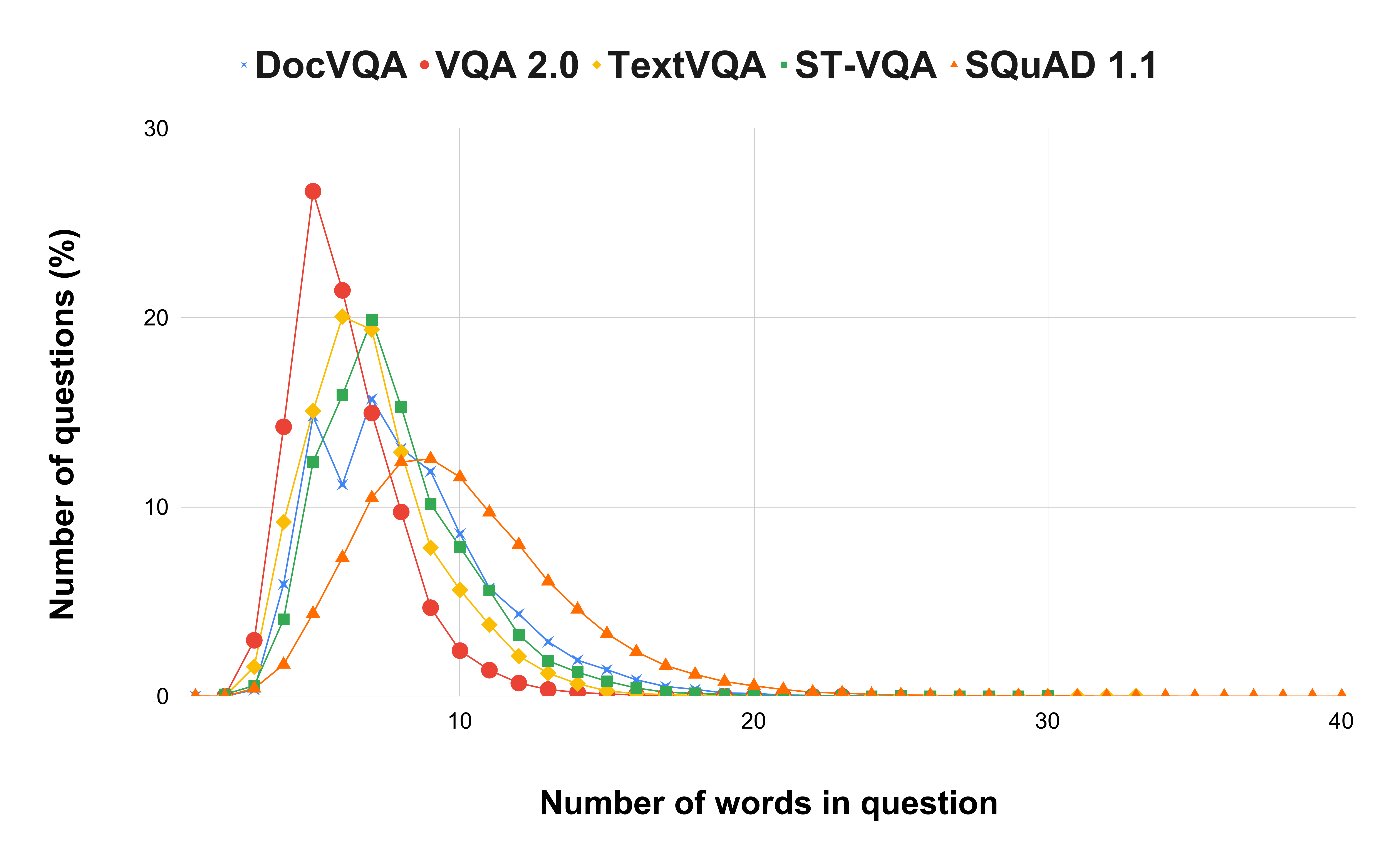}
        \caption{Questions with a particular length.}
        \label{fig:compare_question_lengths}
    \end{subfigure}
    \hspace{1mm}
    \begin{subfigure}[t]{0.32\textwidth}
        \includegraphics[width=1\linewidth]{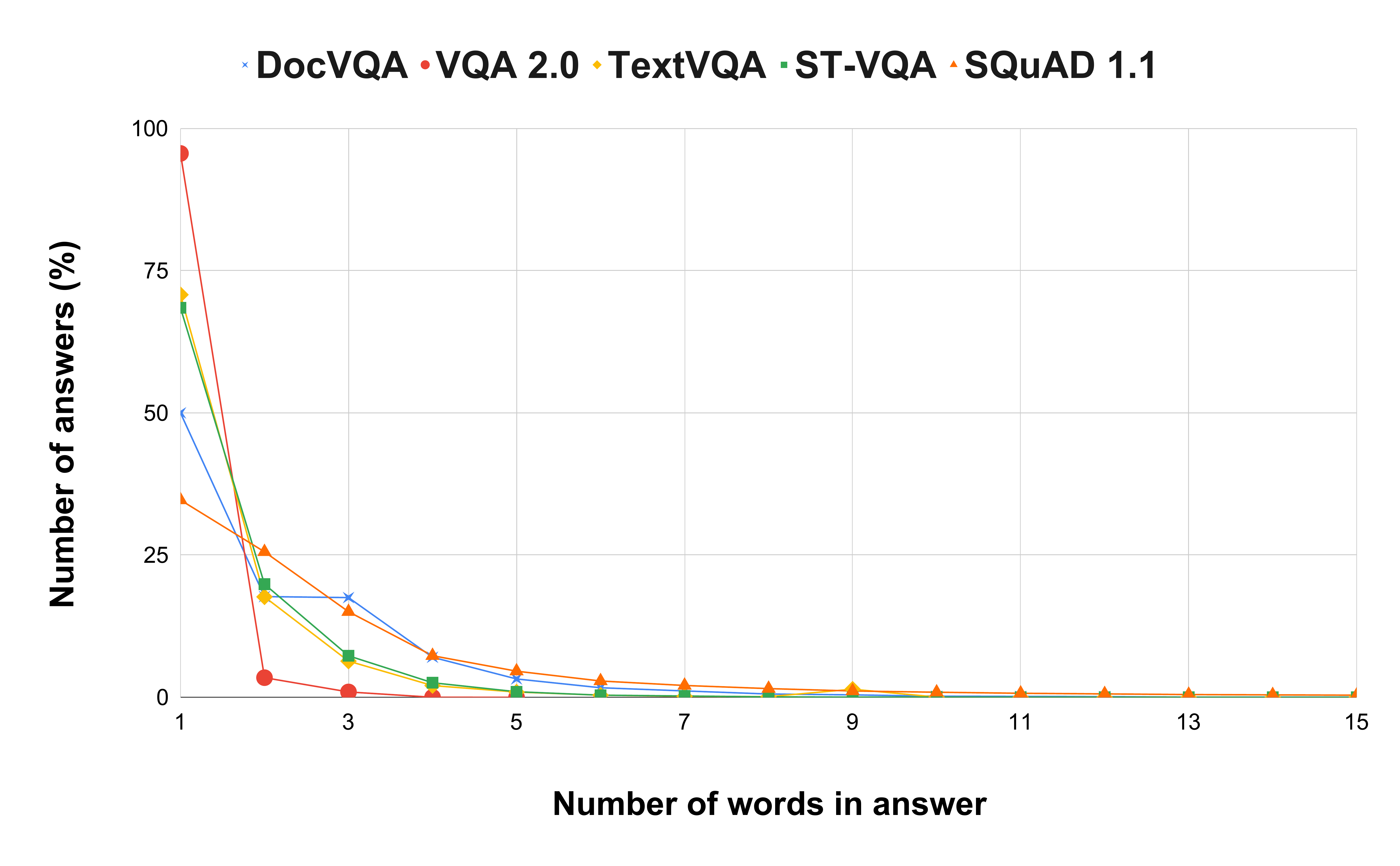}
        \caption{Answers with a particular length.}
        \label{fig:compare_answer_lengths}
    \end{subfigure}
    \hspace{1mm}
    \begin{subfigure}[t]{0.32\textwidth}
        \includegraphics[width=1\linewidth]{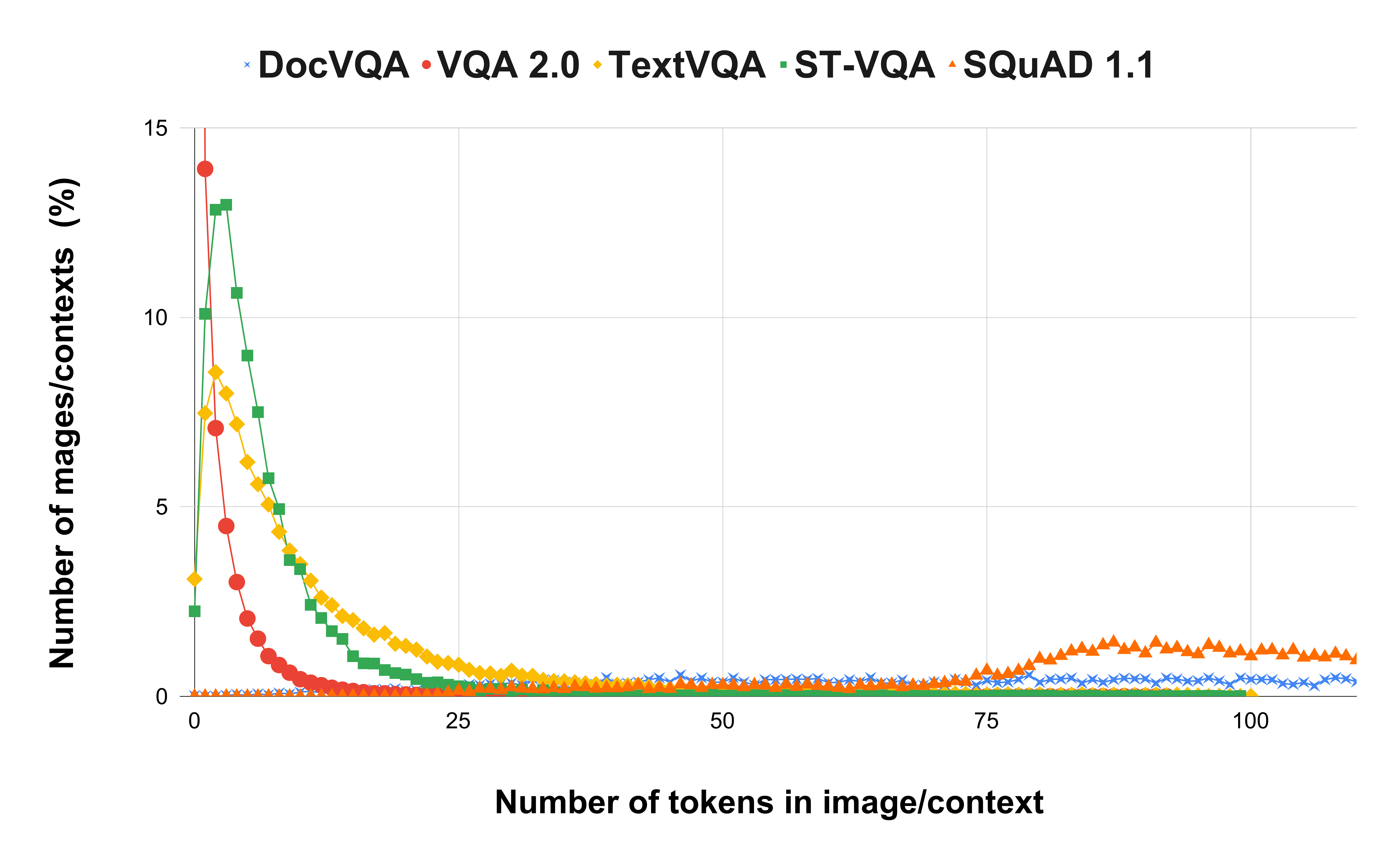}
        \caption{Images/contexts with a particular length}
        \label{fig:compare_document_lengths}
    \end{subfigure}
    \vspace{-2mm}
    \caption{Question, answer and OCR tokens' statistics compared to similar datasets from VQA --- \vqa~\cite{vqa2}, \stvqa~\cite{stvqa_iccv} and \textvqa~\cite{textvqa} and \squad~\cite{squad} reading comprehension dataset. }
\end{figure*}
The second annotation stage aims to verify the data collected in the first stage. 
Here a worker was shown an image and questions defined on it in the first stage (but not the answers from the first stage), and was required to enter answers for the questions.
In this stage 
workers were also required to assign one or more question types to each question. 
The different question types in \datasetName are discussed in~\autoref{sec:stats_analysis}.
During the second stage, if the worker finds a question inapt owing to language issues or ambiguity, an option to flag the question was provided. Such questions are not included in the dataset.

If none of the answers entered in the first stage match exactly with any of the  answers from the second stage, the particular question is sent for review in a third stage. Here questions and answers are editable and the reviewer either accepts the question-answer (after editing if necessary) or ignores it. The third stage review is done by the authors themselves.

\subsection{Statistics and Analysis}
\label{sec:stats_analysis}
The \datasetName comprises $50,000$ questions framed on $12,767$ images. The data is split randomly in an $80-10-10$ ratio to train, validation and test splits. The train split has $39,463$ questions and $10,194$ images, the validation split has $5,349$ questions and $1,286$ images and the test split has $5,188$ questions and $1,287$ images.

As mentioned before, questions are tagged with question type(s) during the second stage of the annotation process.
~\autoref{fig:question_types} shows the 9 question types and percentage of questions under each type. A question type signifies the type of data where the question is grounded.
For example, `table/list' is assigned if answering the question requires understanding of  a table or a list. If the information is in the form of a key:value, the `form' type is assigned. `Layout' is assigned for questions which  require spatial/layout information to find the answer. For example, questions asking for a title or heading, require one to understand structure of the document.
If answer for a question is based on information in the form of sentences/paragraphs type assigned is `running text'. For all questions where answer is based on handwritten text, `handwritten' type is assigned. Note that a question can have more than one type associated with it. (Examples from \datasetName for each question type are given in the supplementary.)

In the following analysis we compare statistics of questions, answers and OCR tokens with other similar datasets for VQA --- \vqa~\cite{vqa2}, \textvqa~\cite{textvqa} and \stvqa~\cite{stvqa_iccv} and \squad~\cite{squad} dataset for reading comprehension. 
Statistics for other datasets are computed based on their publicly available data splits. 
For statistics on OCR tokens, for \datasetName we use OCR tokens generated by a commercial OCR solution. For \vqa, \textvqa and \stvqa we use OCR tokens made available by authors of \lorra~\cite{textvqa} and \mc~\cite{m4c} as part of the MMF~\cite{mmf} framework.

\begin{figure}[b]
    \centering
    \includegraphics[width=0.5\linewidth]{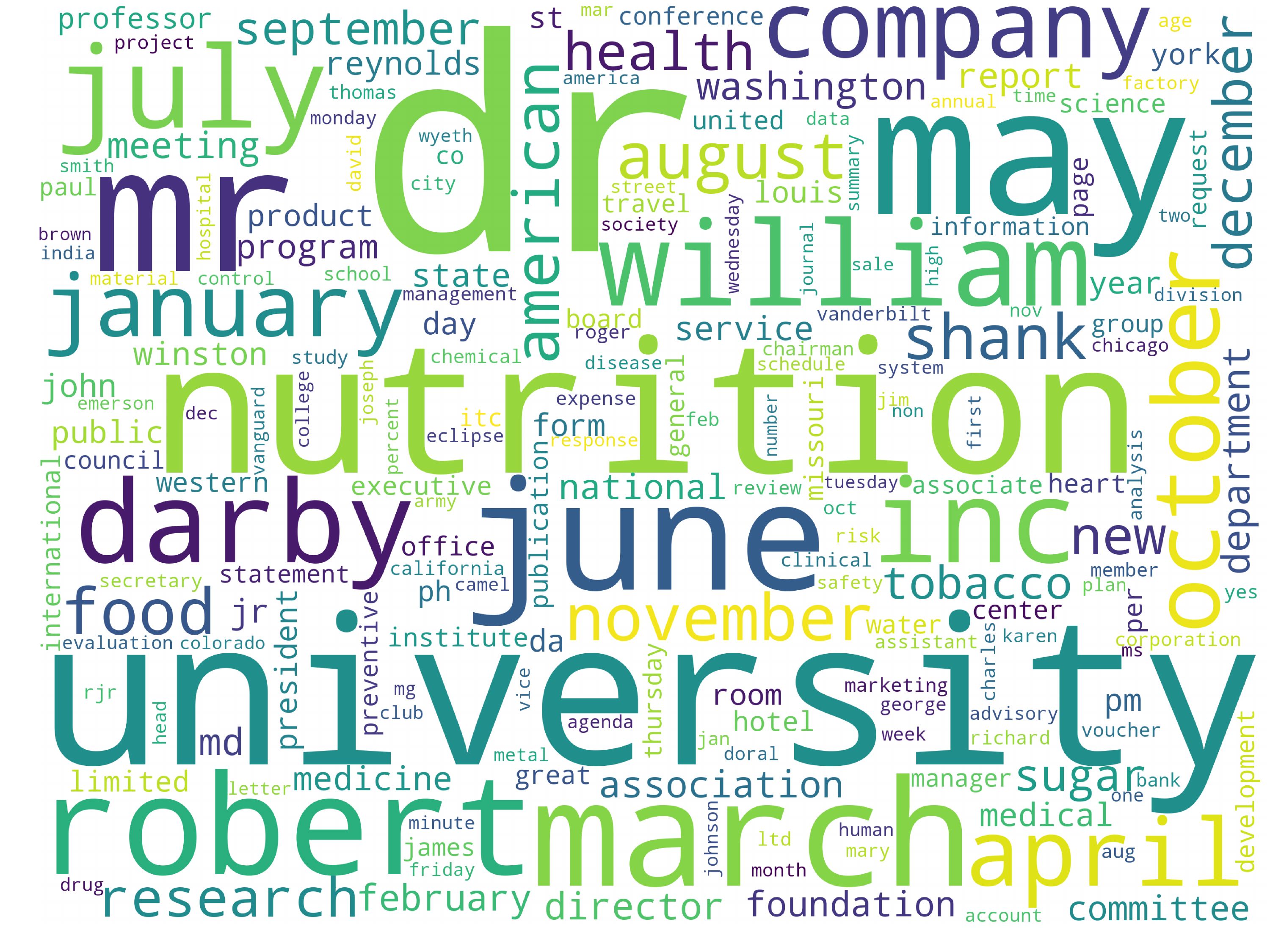}
    \includegraphics[width=0.49\linewidth]{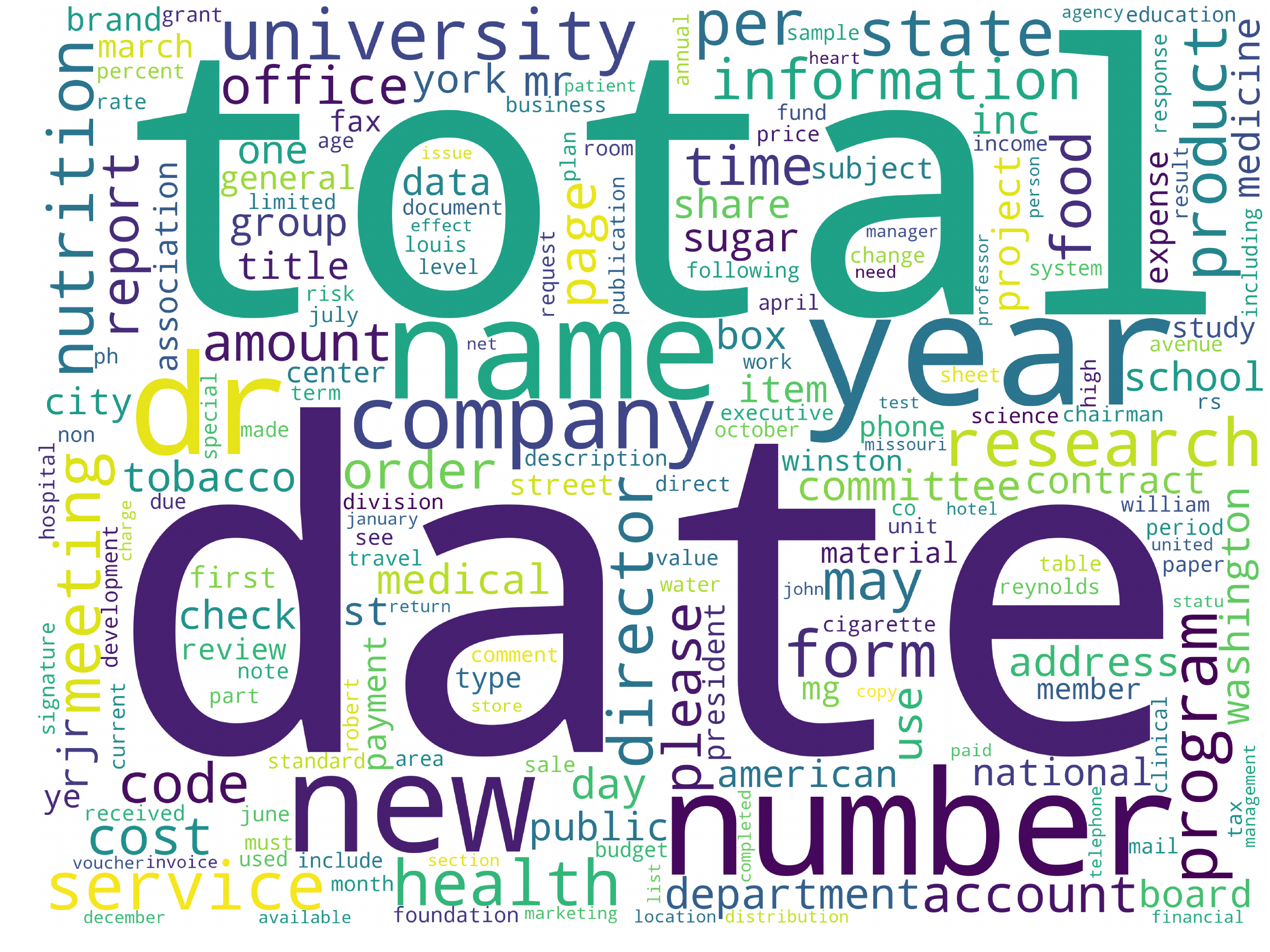}
    \caption{Word clouds of words in answers (left) and words spotted on the document images in the dataset (right)}
    
    \label{fig:wordcloud}
    \vspace{-3mm}

\end{figure}

\begin{figure}[t]
    \centering
    \includegraphics[width=\linewidth]{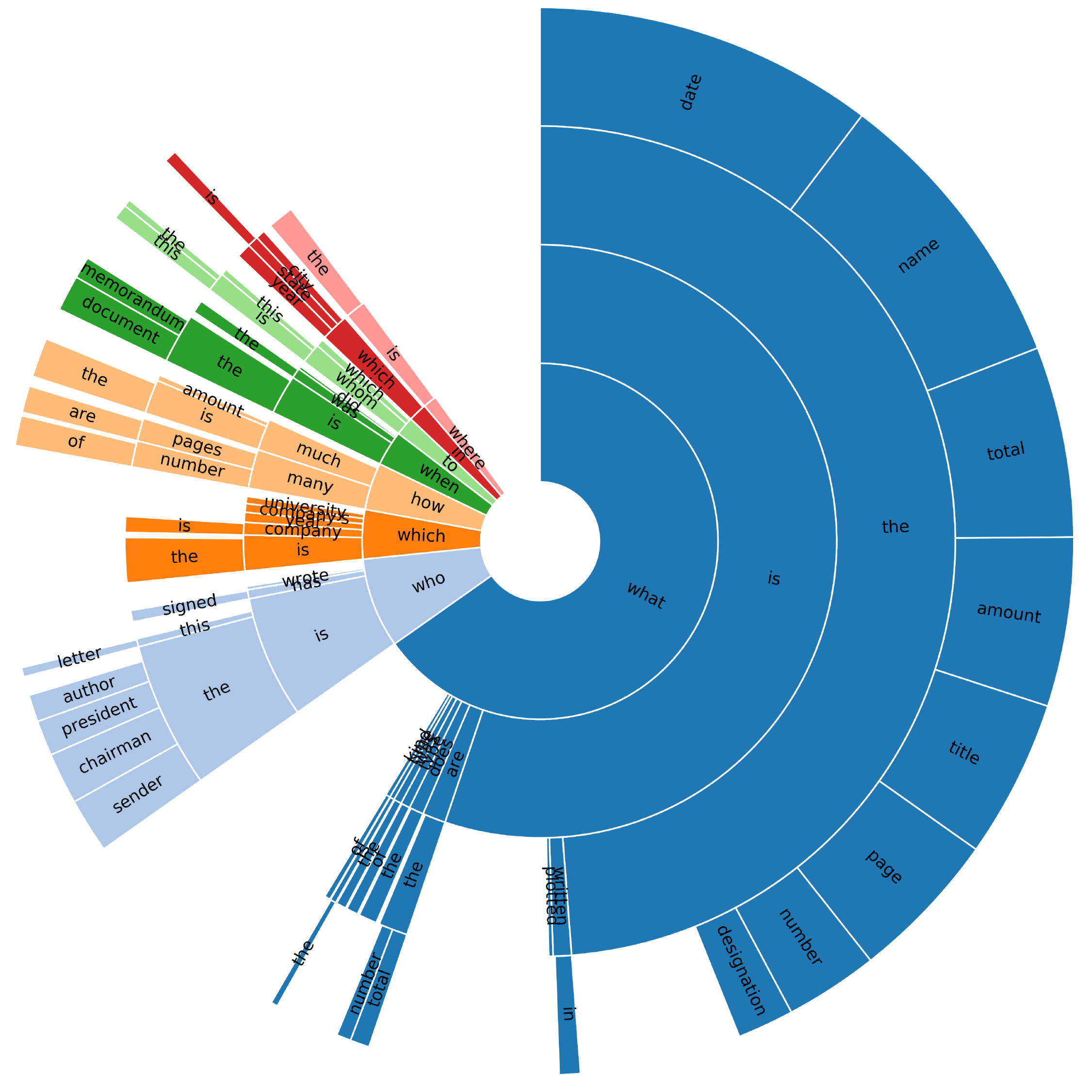}
    \caption{Distribution of questions by  their starting 4-grams. Most questions aim to retrieve common data points in documents such as date, title, total mount and page number. }
    \label{fig:sunburst_4grams}
    \vspace{-5mm}

\end{figure}

\autoref{fig:compare_question_lengths} shows distribution of question lengths for questions in \datasetName compared to other similar datasets.
The average question length is is $8.12$, which is second highest among the compared datasets.
.
In \datasetName $35,362$ ($70.72\%$) questions are unique. 
~\autoref{fig:top_questions} shows the top $15$ most frequent questions and their frequencies. There are questions repeatedly being asked about dates, titles and page numbers. 
A sunburst of first 4 words of the questions is shown in~\autoref{fig:sunburst_4grams}.

It can be seen that a large majority of questions start with ``what is the'', asking for date, title, total, amount or name.

Distribution of answer lengths is shown in~\autoref{fig:compare_answer_lengths}. We observe in the figure that both \datasetName and \squad have a higher number of longer answers compared to the VQA datasets.
The average answer length is $2.17$. 

$63.2\%$ of the answers are unique , which is second only to \squad $(72.5\%)$. The top $15$ answers in the dataset are shown in~\autoref{fig:top_anwers}. 

We observe that almost all of the top answers are numeric values, which is expected since there are a good number of document images of reports and invoices. In~\autoref{fig:top_non_numeric_Answers} we show the top $15$ non numeric answers. These include named entities such as names of people, institutions and places.
The word cloud on the left in~\autoref{fig:wordcloud} shows frequent words in answers. 
Most common words are names of people and names of calendar months.

In~\autoref{fig:compare_document_lengths} we show the number of images (or `context's in case of \squad) containing a particular number of text tokens. 
Average number of text tokens in an image or context is the highest in the case of \datasetName ($182.75$). It is considerably  higher compared to \squad where contexts are usually small paragraphs whose average length is $117.23$. In case of VQA datasets which comprise real world images average number of OCR tokens is not more than $13$. Word cloud on the right in~\autoref{fig:wordcloud} shows the most common words spotted by the OCR on the images in \datasetName. We observe that there is high overlap between common OCR tokens and words in answers.


\section{Baselines}
\label{sec:baselines}
In  this section we  explain the baselines we use, including heuristics and trained models. 

\subsection{Heuristics and Upper Bounds}
\label{sec:heuristics}
Heuristics we evaluate are: (i) \textbf{Random answer:} measures performance when we pick a random answer from the answers in the train split. (ii) \textbf{Random OCR token:} performance when a random OCR token from the given document image is picked as the answer. (iii) \textbf{Longest OCR token} is the case when the longest OCR token in the given document is selected as the answer. (iv) \textbf{Majority answer} measures the performance when the most frequent answer in the train split is considered as the answer.

We also compute following upper bounds: (i) \textbf{Vocab UB:} This upper bound measures performance upper bound one can get by predicting correct answers for the questions, provided the correct answer is present in a vocabulary of answers, comprising all answers which occur more than once in the train split.
(ii) \textbf{OCR substring UB:} is the upper bound on predicting the correct answer provided the answer can be found as a substring in the sequence of OCR tokens.
The  sequence is made by serializing the OCR tokens recognized in the documents as a sequence separated by space, in top-left to bottom-right order.
(iii) \textbf{OCR subsequence UB:} upper bound on predicting the correct answer, provided the answer is a subsequence of the OCR tokens' sequence.

\subsection{VQA Models}
\label{sec: vqa models}
For evaluating performance of existing VQA models on \datasetName we employ two models which take the text present in the images into consideration while answering questions -- Look, Read, Reason \& Answer (\lorra)~\cite{textvqa} and Multimodal Multi-Copy Mesh(\mc)~\cite{m4c}.

\textbf{\lorra:} follows a bottom-up and top-down attention~\cite{topdown_bottomup} scheme with additional bottom-up attention over OCR tokens from the images.
In \lorra, tokens in a question are first embedded using a pre-trained embedding (GloVe~\cite{glove}) and then these tokens are iteratively encoded using an LSTM~\cite{lstm} encoder. The model uses two types of spatial features to represent the visual  information from the images - (i) grid convolutional features from a  Resnet-152~\cite{resnet} which is pre-trained on ImageNet~\cite{imagenet} and (ii) features extracted from  bounding box proposals from a Faster R-CNN~\cite{faster-r-cnn} object detection model, pre-trained on Visual Genome~\cite{visual_genome}. OCR tokens from the image are embedded using a pre-trained word embedding (FastText~\cite{fasttext}). An attention mechanism is used to compute an attention weighed average of the image features as well the OCR tokens' embeddings. These averaged features are combined and fed into an output module. The classification layer of the model, predicts an answer either from a fixed vocabulary (made from answers in train set) or copy an answer from a dynamic vocabulary which essentially is the list of OCR tokens in an image. Here the copy mechanism can copy only one of the OCR tokens from the image. Consequently it cannot output an answer which is a combination of two or more OCR tokens.

\textbf{M4C}: uses a multimodal transformer and an iterative answer prediction module. 
Here tokens in questions are embedded using a BERT model~\cite{bert}. Images are represented using (i) appearance features of the objects detected using a Faster-RCNN pretrained on Visual Genome~\cite{visual_genome} and  (ii) location information - bounding box coordinates of the detected objects. Each OCR token recognized from the image is represented using (i) a pretrained word embedding (FastText), (ii) appearance feature of the token's bounding box from the same Faster R-CNN which is used for appearance features of objects (iii) PHOC~\cite{phoc} representation of the token and (iv) bounding box coordinates of the token. Then these feature representations of the three entities (question tokens, objects and OCR tokens) are projected to a common, learned embedding space. 
Then a stack of Transformer~\cite{attention_is_all_you_need} layers are applied over these features in the common embedding space. 
The multi-head self attention in transformers enable both inter-entity and intra-entity attention. Finally, answers are predicted through iterative decoding in an auto-regressive manner. Here the fixed vocabulary used for the closed answer space is made up of the most common answer words in the train split. Note that in this case the fixed vocabulary comprises of answer words, not answers itself as in the case of \lorra.
At each step in the decoding, the decoded word is either an OCR token from the image or a word from the fixed vocabulary of common answer words.

In our experiments we use original \lorra  and \mc models and few variants of these models.
Document images in \datasetName usually contain higher number of text tokens compared to images in scene text VQA datasets. Hence we try out larger dynamic vocabularies (i.e. more OCR tokens are considered from the images) for both \lorra and \mc. For both the models we also evaluate performance when no fixed vocabulary is used.

Since the notion of visual objects in real word images is not directly applicable in case of document images, we also try out variants of \lorra and \mc where features of objects are omitted.

\subsection{Reading Comprehension Models}
\label{sec:RC_models}
In addition to the VQA models which can read text, we try out extractive question answering / reading comprehension models from NLP. In particular, we use BERT~\cite{bert} question answering models. BERT is a method of pre-training language representations from unlabelled text using transformers~\cite{attention_is_all_you_need}. These pretrained models can then be used for downstream tasks with just an additional output layer. In the case of extractive Question Answering, this is an output layer to predict start and end indices of the answer span.

\section{Experiments}
\label{sec:experiments}
In this section we explain evaluation metrics and our  experimental settings and report results of experiments.
\subsection{Evaluation Metrics}
\label{sec:evaluation}
Two evaluation metrics we use are Average Normalized Levenshtein Similarity (ANLS) and Accuracy (Acc.).
ANLS was originally proposed for evaluation of VQA on \stvqa~\cite{st-vqa_challenge}.
The Accuracy metric measures percentage of questions for which the predicted answer matches exactly with any of the target answers for the question.
Accuracy metric  awards a zero score even when the prediction is only a little different from the target answer. Since no OCR is perfect, we propose to use ANLS as our primary evaluation metric, so that minor answer mismatches stemming from OCR errors are not severely penalized.

\begin{table}[t]
\small
\begin{tabularx}{\linewidth}{X |c c c c}
\toprule
&  \multicolumn{2}{c}{val} & \multicolumn{2}{c}{test} \\             
  Baseline   & ANLS & Acc.  & ANLS & Acc.  \\
\midrule

Human   & - & - & 0.981 & 94.36 \\ 
Random answer   & 0.003 & 0.00 & 0.003 & 0.00 \\  
Rnadom OCR token   & 0.013 & 0.52 & 0.014 & 0.58 \\  
Longest OCR token     & 0.002 & 0.05 & 0.003 & 0.07 \\  
Majority answer  & 0.017 & 0.90 & 0.017 & 0.89 \\  
Vocab UB   & - & 31.31 & - & 33.78 \\  

OCR substring UB & - & 85.64 & - & 87.00 \\  
OCR subsequence UB & - & 76.37 & - & 77.00 \\ 
\bottomrule
\end{tabularx}
\caption{Evaluation of different heuristics and upper bounds. Predicting random answers or majority answer do not even yield 1\% accuracy. Answers are a substring of the serialized OCR output in more than 85\% of the cases.}
\label{tab:human_heuristics}
\vspace{-2mm}

\end{table}


\begin{table*}[h]
\small
\begin{tabularx}{\linewidth}{X | c c c | c c c c}
\toprule
 &   &    &  & \multicolumn{2}{c}{val} & \multicolumn{2}{c}{test} \\             
Method  & Objects' feature & Fixed vocab.  & Dynamic vocab. size   & ANLS & Acc.  & ANLS & Acc.  \\
\midrule
\multirow{4}{*}{\lorra~\cite{textvqa}}
 & \cmark & \cmark & 50 & \textbf{0.110} & 7.22 & \textbf{0.112} & 7.63\\ 

 & \cmark & \xmark & 50 & 0.041 & 2.64 & 0.037 & 2.58\\ 
& \xmark & \cmark & 50 & 0.102 & 6.73 & 0.100 & 6.43\\ 
& \cmark & \cmark & 150 & 0.101 & 7.09 & 0.102 & 7.22\\  
 & \cmark & \cmark & 500 & 0.094 & 6.41 & 0.095 & 6.31\\  

\midrule

\multirow{4}{*}{\mc~\cite{m4c}}
 & \cmark & \cmark & 50 & 0.292 & 18.34 & 0.306 & 18.75\\ 
& \cmark & \xmark & 50 & 0.216 & 12.44 & 0.219 & 12.15\\ 
 & \xmark & \cmark & 50 & 0.294 & 18.75 & 0.310 & 18.92\\ 
 & \xmark & \cmark & 150 & 0.352 & 22.66 & 0.360 & 22.35\\ 
& \xmark & \cmark & 300 & 0.367 & 23.99 & 0.375 & 23.90\\ 
& \xmark & \cmark & 500 & \textbf{0.385} & 24.73 & \textbf{0.391} & 24.81\\ 
    
\bottomrule
\end{tabularx}
\caption{Performance of the VQA models which are capable of reading text --- \lorra~\cite{textvqa} and \mc\cite{m4c}. Detection of visual objects and their features (bottom-up attention), which is a common practice in VQA is ineffective in case of \datasetName.
}
\label{tab:vqa_results}
\vspace{-2mm}

\end{table*}

\subsection{Experimental setup}
\label{sec: experimental setup}
For measuring  human performance , we collect answers for all questions in test split, with help a few volunteers from our institution.

In all our experiments including heuristics and trained baselines, OCR tokens we use are extracted using a commercial OCR application.
For the heuristics and upper bounds we use a vocabulary $4,341$ answers which occur more than once in the train split.

For \lorra and \mc models we use official implementations available as part of the MMF framework~\cite{mmf}. The training settings and hyper parameters are same as the ones reported in the original works. The fixed vocabulary we use for \lorra is same as the vocabulary we use for computing vocabulary based heuristics and upper bounds. For \mc the fixed vocabulary we use is  a vocabulary of the $5,000$ most frequent words from the answers in the train split.

For QA using BERT, three pre-trained BERT models\footnote{\url{https://huggingface.co/transformers/pretrained_models.html}} from the Transformers library~\cite{huggingface} are used. The models we use are bert-base-uncased, bert-large-uncased-whole-word-masking and bert-large-uncased-whole-word-masking-finetuned-squad. We abbreviate the model names as bert-base, bert-large and bert-large-squad respectively.
Among these, bert-large-squad is a  pre-trained model which is also finetuned on \squad for question answering. 
In case of extractive question answering or reading comprehension datasets `contexts' on which questions are asked are passages of electronic text. But in \datasetName  `contexts' are document images.
Hence to finetune the BERT QA models on \datasetName we need to prepare the data in SQuAD style format where the answer to a question is a `span' of the context, defined by start and end indices of the answer. To this end we first serialize the OCR tokens recognized on the document images to a single string, separated by space, in top-left to bottom-right order. To approximate the answer spans we follow an approach proposed in TriviaQA~\cite{triviaqa}, which is to find the first match of the answer string in the serialized OCR string.

The bert-base model is finetuned on \datasetName on 2 Nvidia GeForce 1080 Ti GPUs, for 2 epochs, with a batch size of 32.
We use Adam optimizer~\cite{adam} with a learning rate of $5e-05$.
The bert-large and bert-large-squad models are finetuned on 4 GPUs for 6 epochs with a batch size of 8, and a learning rate of $2e-05$.

\begin{table}[b]
\small
\begin{tabularx}{\linewidth}{X |X| c c c c}
\toprule
&   &\multicolumn{2}{c}{val} & \multicolumn{2}{c}{test} \\             
Pretrained model   & \datasetName finetune  & ANLS & Acc.  & ANLS & Acc.  \\
\midrule

bert-base & \cmark  & 0.556 & 45.6 & 0.574 & 47.6 \\ 
bert-large- & \cmark  & 0.594 & 49.28 & 0.610 & 51.08 \\ 
bert-large-squad  & \xmark  & 0.462 & 36.72 & 0.475 & 38.26 \\ 
bert-large-squad   & \cmark  & \textbf{0.655} & 54.48 & \textbf{0.665} & 55.77 \\ 

\bottomrule
\end{tabularx}
\caption{Performance of BERT question answering models. A BERT\textsubscript{LARGE}  model which is fine tuned on both \squad~\cite{squad} and \datasetName performs the best.}
\label{tab:bert_results}
\vspace{-2mm}

\end{table}

\subsection{Results}
\label{sec:results}
Results of all heuristic approaches and upper bounds are reported in~\autoref{tab:human_heuristics}.
We can see that none of the heuristics get even a  $1\%$ accuracy on the validation or test splits.
\begin{figure*}[h]
\begin{center}
\begin{tabular}{p{0.32\linewidth}p{0.32\linewidth}p{0.32\linewidth}}
    \includegraphics[width=\linewidth,height=1\linewidth]{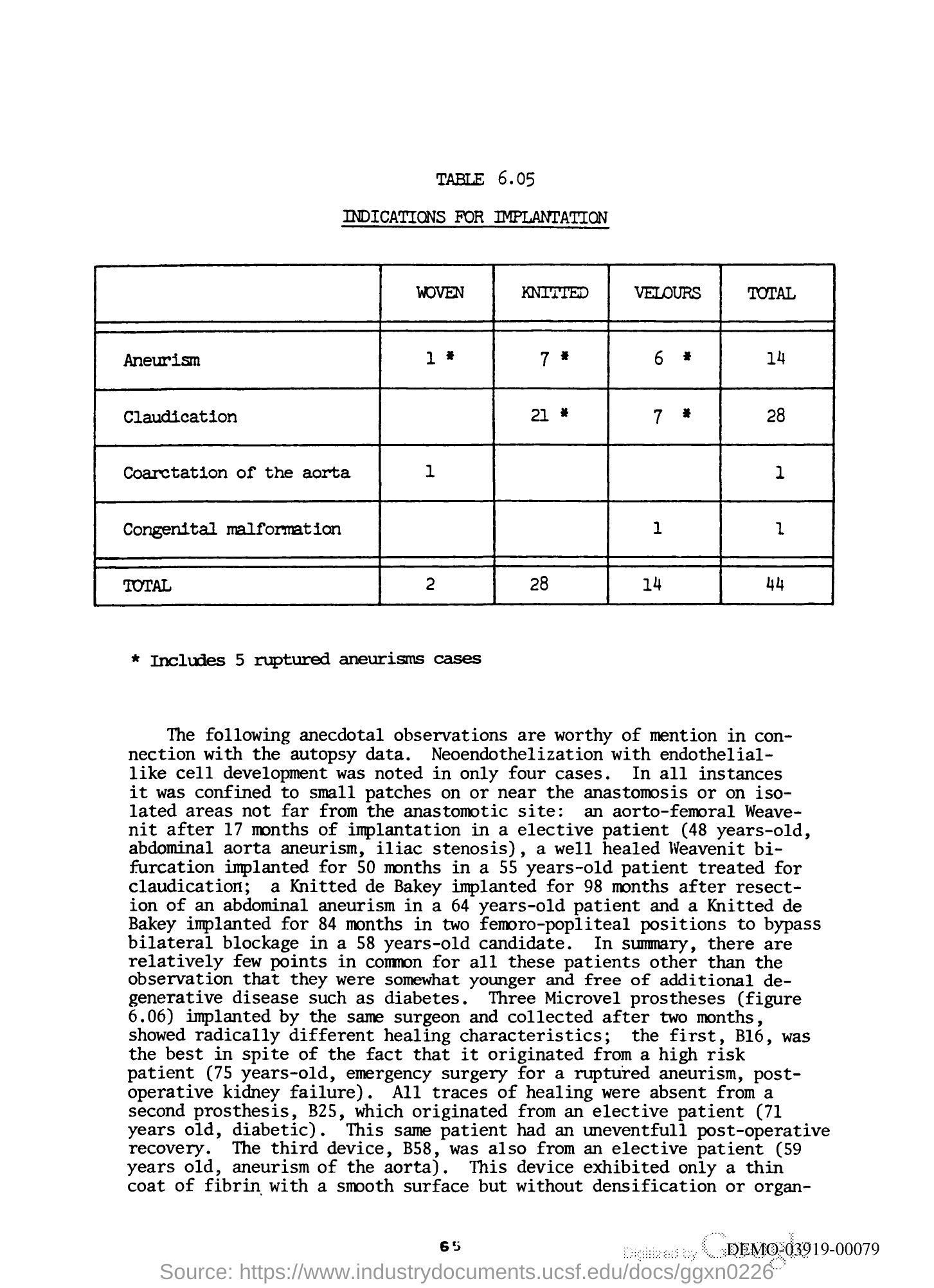} &
    \includegraphics[width=\linewidth,height=1\linewidth]{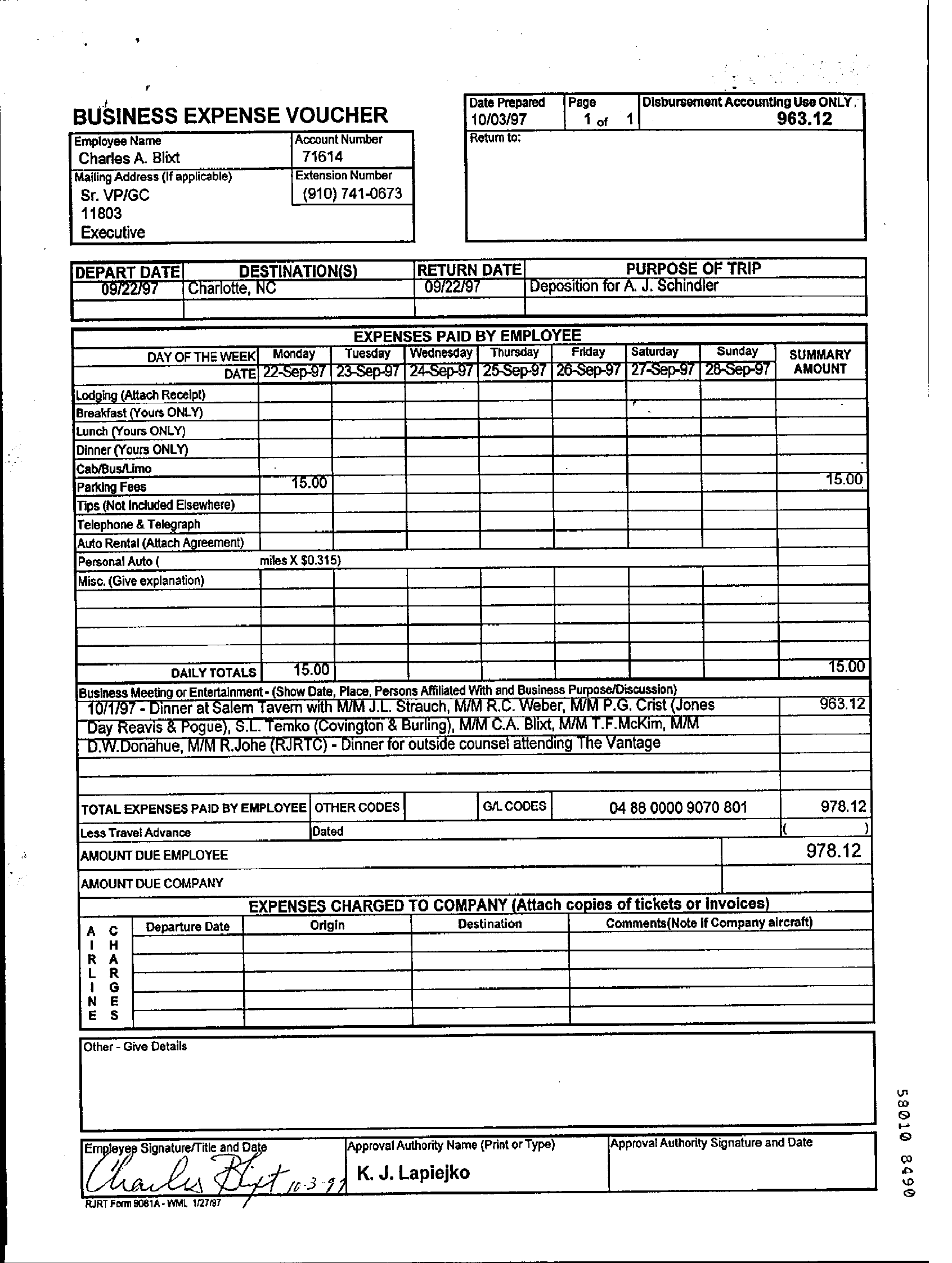} &
    \includegraphics[width=\linewidth,height=1\linewidth]{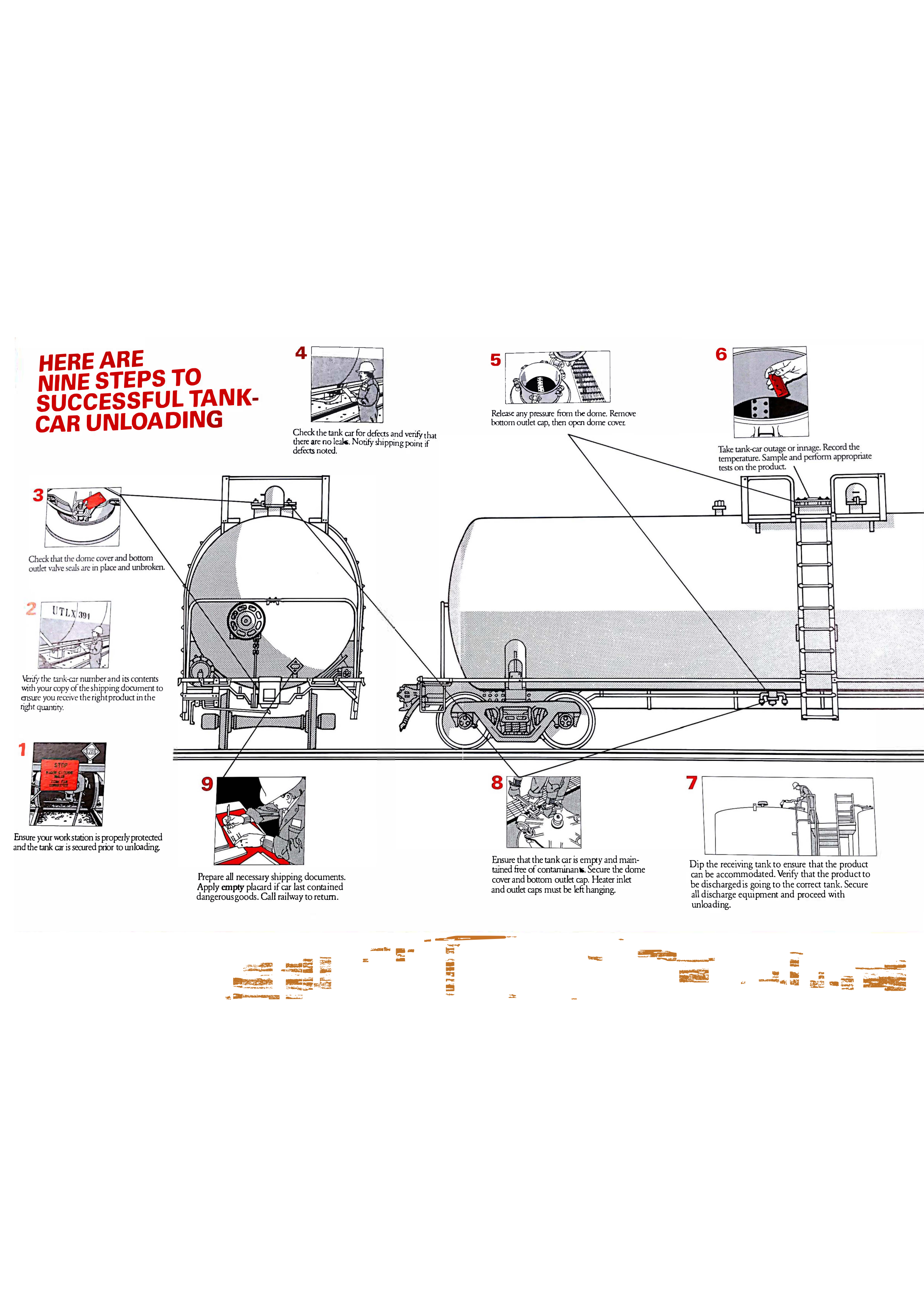}

    \\

  \fbox{\includegraphics[width=\linewidth,height=0.3\linewidth]{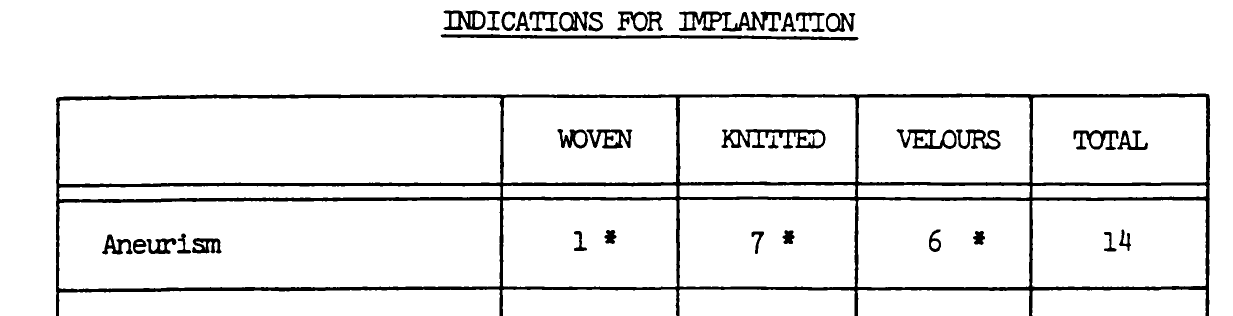}} &
  \fbox{\includegraphics[width=\linewidth,height=0.3\linewidth]{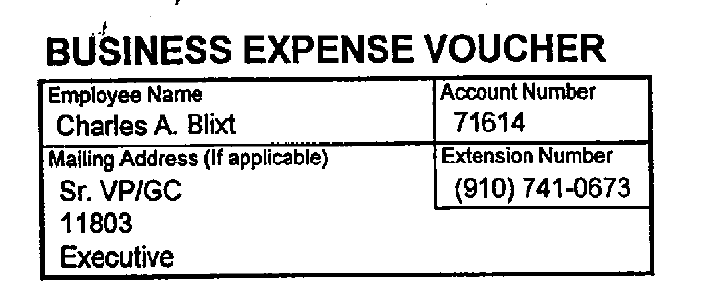}} &
 \fbox{\includegraphics[width=\linewidth,height=0.3\linewidth]{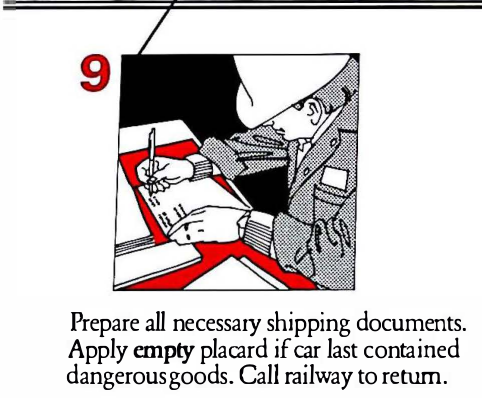}}

    \\
    \footnotesize{\fontfamily{qhv}\selectfont \textbf{Q:} What is the underlined heading just above the table? } \par \footnotesize{\fontfamily{qhv}\selectfont \textbf{GT:} \color{blue}Indications for implantation} 
    \par \footnotesize{\fontfamily{qhv}\selectfont \textbf{M4C best:} \color{green}indications for implantation} 
     \par \footnotesize{\fontfamily{qhv}\selectfont \textbf{BERT best:} \color{red}total aneurism} 
     \par \footnotesize{\fontfamily{qhv}\selectfont \textbf{Human:} \color{green}indications for implantation} 
    &

    \footnotesize{\fontfamily{qhv}\selectfont \textbf{Q:} What is the Extension Number as per the voucher? } \par  \footnotesize{\fontfamily{qhv}\selectfont \textbf{GT:} \color{blue}(910) 741-0673} 
    \par \footnotesize{\fontfamily{qhv}\selectfont \textbf{M4C best:} \color{red} 963.12} 
     \par \footnotesize{\fontfamily{qhv}\selectfont \textbf{BERT best:} \color{green}(910) 741-0673} 
     \par \footnotesize{\fontfamily{qhv}\selectfont \textbf{Human:} \color{green}(910) 741-0673} 
    &
    
  \footnotesize{\fontfamily{qhv}\selectfont \textbf{Q:} How many boxed illustrations are there ? } \par      
  
  \footnotesize{\fontfamily{qhv}\selectfont \textbf{GT:} \color{blue} 9} 
    \par \footnotesize{\fontfamily{qhv}\selectfont \textbf{M4C best:} \color{red} 4} 
     \par \footnotesize{\fontfamily{qhv}\selectfont \textbf{BERT best:} \color{red} 4} 
     \par \footnotesize{\fontfamily{qhv}\selectfont \textbf{Human:} \color{green} 9} 
   
\end{tabular}
\end{center}
\caption{Qualitative results from our experiments. The leftmost example is a `layout' type question answered correctly by the \mc model but erred by the BERT model. In the second example the BERT model correctly answers a question on a form while the \mc model fails. In case of the rightmost example, both models fail to understand a step by step illustration.}
\label{fig:qualitative_results}
\vspace{-4mm}

\end{figure*}

\textit{OCR substring UB} yields  more than $85\%$ accuracy on both validation and test splits. 
It has a downside that the substring match in all cases need not be an actual answer match. For example if the answer is ``2" which is the most common answer in the dataset, it will match with a ``2" in ``2020" or a ``2" in ``2pac''. This is the reason why we evaluate the \textit{OCR subsequence UB}. An answer is a sub sequence of the serialized OCR output for around  $76\%$ of the questions in both validation and test splits.

Results of our trained VQA baselines are shown in~\autoref{tab:vqa_results}. First rows for both the methods report results of the original model proposed by the respective authors.
In case of \lorra the original setting proposed by the authors yields the best results compared to the variants we try out. With no fixed vocabulary, the performance of the model drops sharply suggesting that the model primarily relies on the fixed vocabulary to output answers. Larger dynamic vocabulary results in a slight performance drop suggesting that incorporating more OCR tokens from the document images does little help. 
Unlike \lorra, \mc benefits from a larger dynamic vocabulary. Increasing the size of the dynamic vocabulary from $50$ to $500$ improves the ANLS by around $50\%$. And in case of \mc, the setting where features of objects are omitted, performs slightly better compared to the original setting.

\begin{figure}[h]
    \centering
    \includegraphics[width=\linewidth]{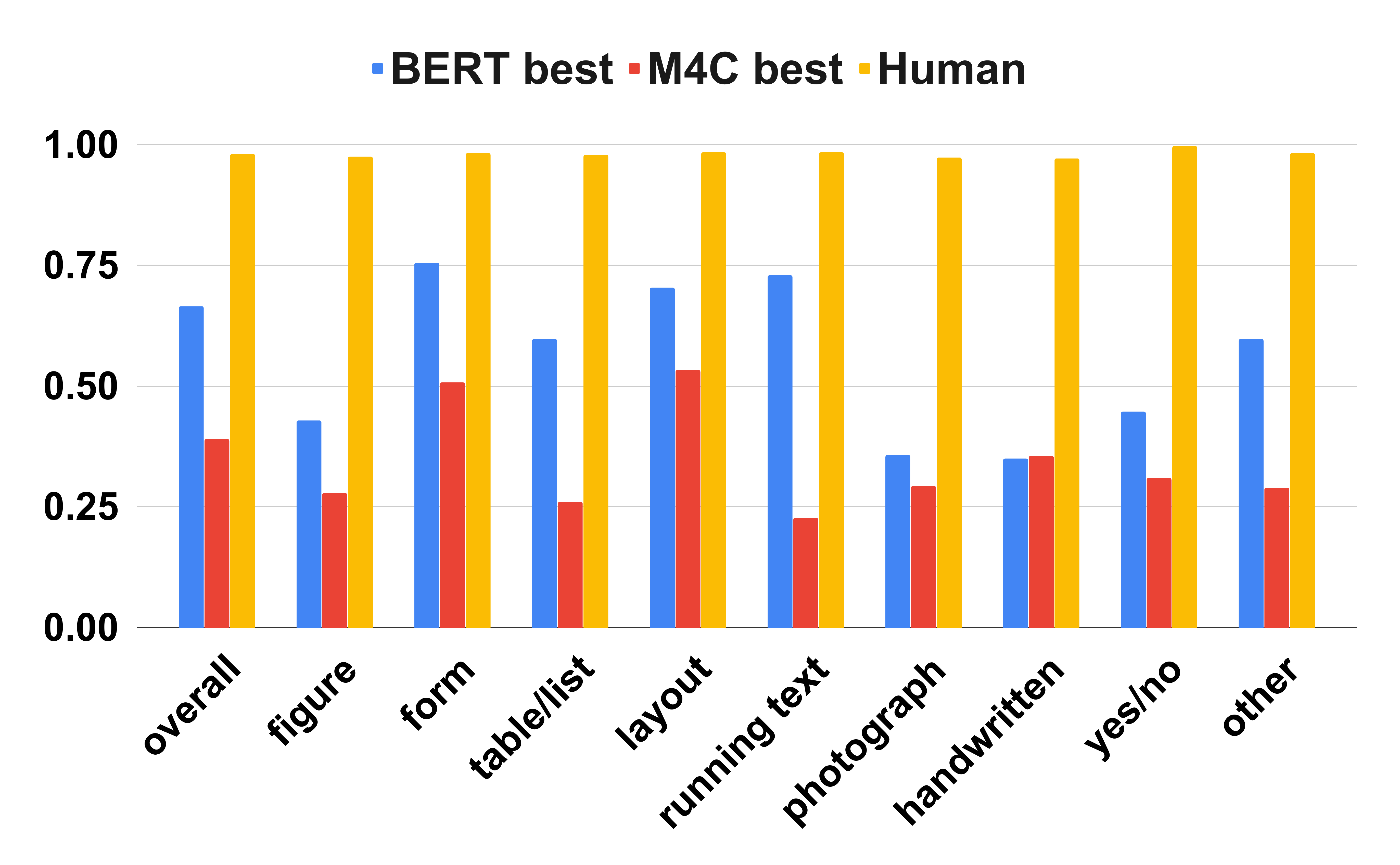}
    \caption{Best baselines from VQA space and reading comprehension space pitted against the human performance for different question types. We need models which can understand figures and text on photographs better. We need better handwriting recognizers too!}
    \label{fig:performance_question_type}
    \vspace{-4mm}

\end{figure}

Results of the BERT question answering models are  reported in~\autoref{tab:bert_results}. We observe that all BERT models perform better than the best VQA baseline using \mc (last row in~\ref{tab:vqa_results}). 
The best performing model out of all the baselines analysed is the bert-large-squad model, finetuned on \datasetName. Answers predicted by this model match one of the target answers exactly for around  $55\%$ of the questions.

In~\autoref{fig:performance_question_type} we show performance by question type. We compare the best models among VQA models and BERT question answering models against the human performance on the test split. We observe that the human performance is uniform while the models' performance vary for different question types.
In~\autoref{fig:qualitative_results} we show a few qualitative results from our experiments.
 
\section{Conclusion}
We introduce a new data set and an associated VQA task with the aim to inspire a ``purpose-driven'' approach in document image analysis and recognition research. Our baselines and initial results motivate simultaneous use of visual and textual cues for answering questions asked on document images. This could drive methods that use the low-level cues (text, layout, arrangements) and high-level goals (purpose, relationship, domain knowledge) in solving problems of practical importance.
\vspace{2mm}

\noindent\textbf{Acknowledgements}

We thank Amazon for supporting the annotation effort, and Dr. R. Manmatha for many useful discussions and inputs. This work is partly supported by MeitY, Government of India, the project TIN2017-89779-P, an Amazon AWS Research Award and the CERCA Programme.

\newpage

{\small
\bibliographystyle{ieee_fullname}
\bibliography{references}
}

\appendix
\counterwithin{figure}{section}
\counterwithin{table}{section}
\counterwithin{equation}{section}

\newpage
\phantom{emptytext}
\newpage

\section{Screen grabs of Annotation Tool}
\label{appendix:screen grabs}

\begin{figure*}[h]
    \centering
    \includegraphics[width=0.8\linewidth]{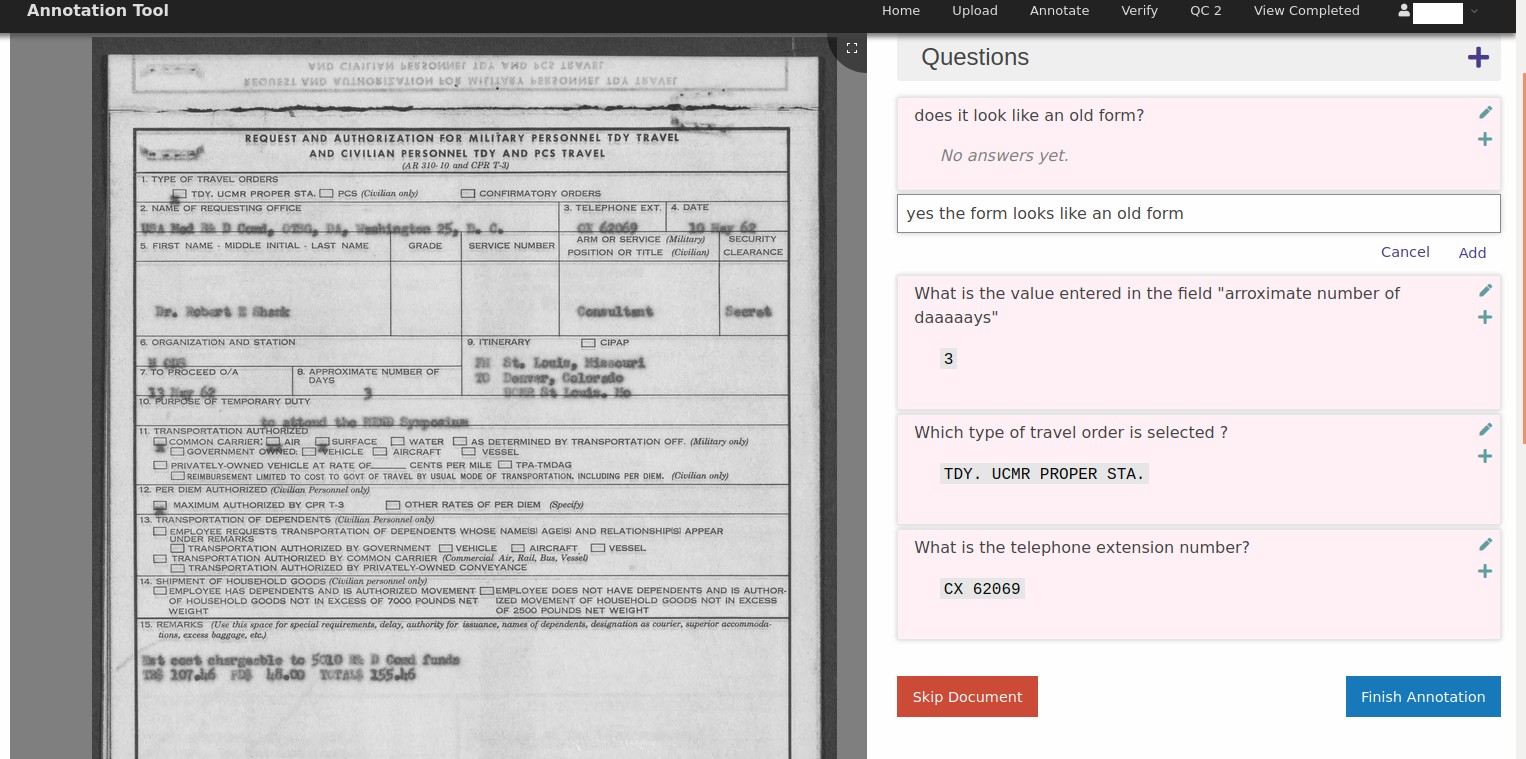}
    \caption{\textbf{Annotation stage 1 - Question Answer Collection: } Questions and answers are collected for a given document image. Annotator can add upto 10 questions for a document. The document can be skipped if it is not possible to frame questions on it.}
    \label{fig:ann_stage1}
\end{figure*}

\begin{figure*}[h]
    \centering
    \includegraphics[width=0.8\linewidth]{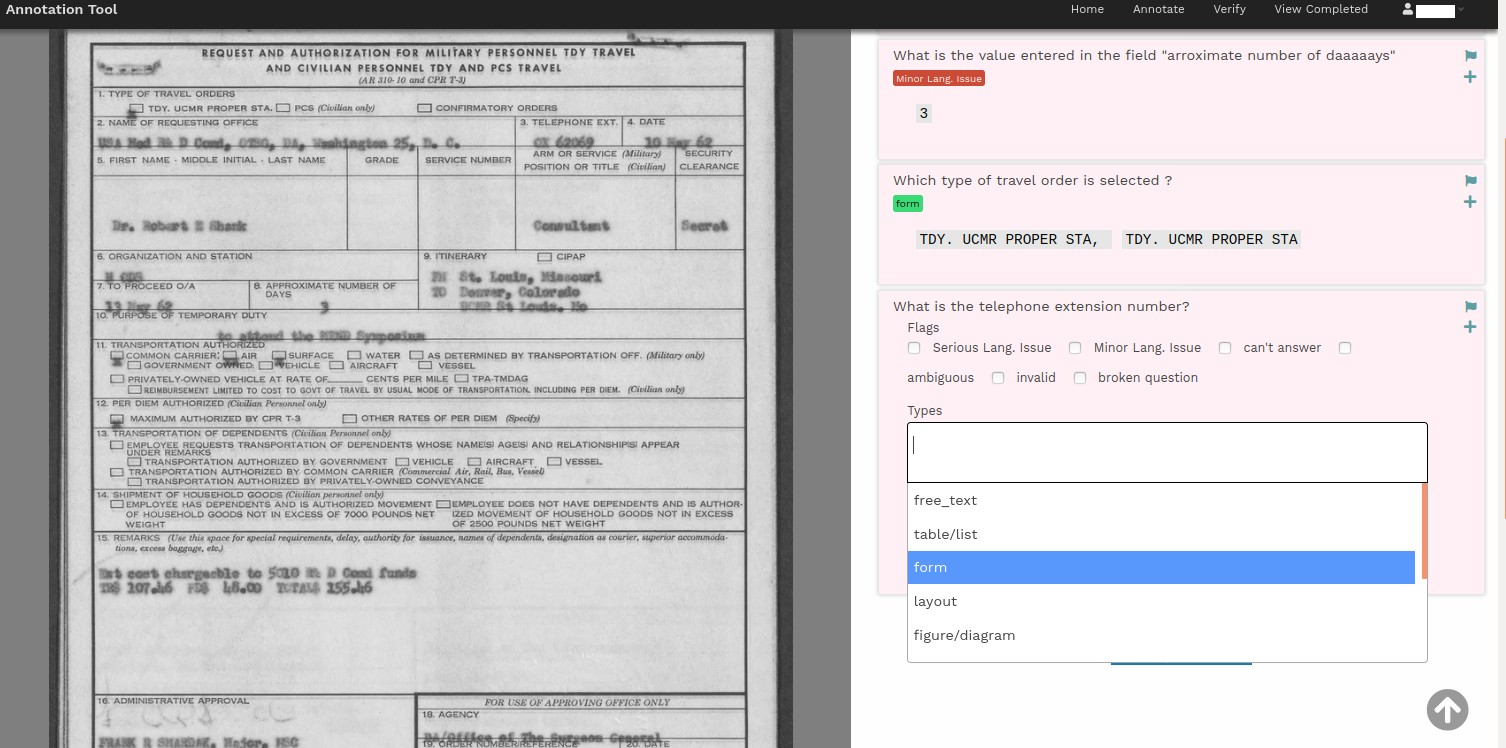}
    \caption{\textbf{Annotation stage 2 - Data Verification: } For each question shown annotators have to  (i) enter answer(s)  (answer(s) from first stage are not shown)  and (ii) Tag the question with one or more question types from the 9 question types shown in a drop-down (question types assigned to a question are shown in green highlight color.)  or  (iii) flag/ignore the question by selecting the  check-box corresponding to one of the reasons such as ``invalid question", ``Serious lang. issue'' etc. ( the reasons chosen for flagging a question are shown in red highlight color )}
    \label{fig:ann_stage2}
\end{figure*}

\begin{figure*}[h]
    \centering
    \includegraphics[width=0.8\linewidth]{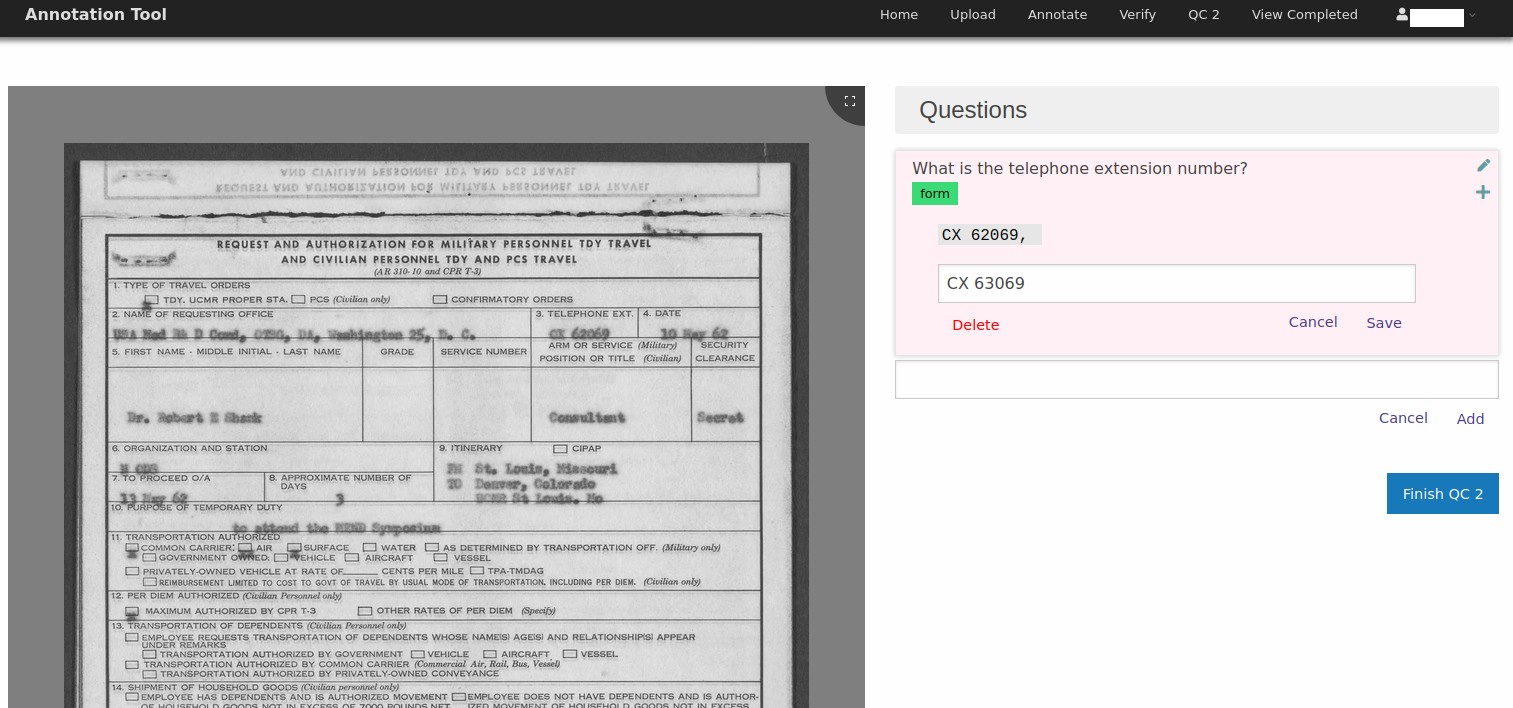}
    \caption{\textbf{Annotation Stage 3 : Reviewing answer mismatch cases : } If none of the answers entered in the first stage for a question  match with any of the answers entered in the second stage, the question is sent for review in a third stage. This review is handled by the authors and reviewer is allowed to edit question as well answers or add new answers before accepting the question.}
    \label{fig:ann_stage3}
\end{figure*}

As mentioned in Section 3.1 in the main paper, annotation process involves three stages.
In ~\autoref{fig:ann_stage1}, ~\autoref{fig:ann_stage2} and~\autoref{fig:ann_stage3} we show screen grabs from stage 1, stage 2 and stage 3 of the annotation process respectively.

\section{Examples of Question Types}
\label{appendix:question_types}
We define  9 question types, based on the kind of reasoning required to answer a question.
Question types are assigned at the second stage of the annotation. We discuss the question types in Section 3.2. in the main paper. 

\begin{figure*}[h]
\begin{center}
\begin{tabular}{p{0.4\linewidth}p{0.44\linewidth}}
\multicolumn{2}{c}{\includegraphics[width=0.6\linewidth]{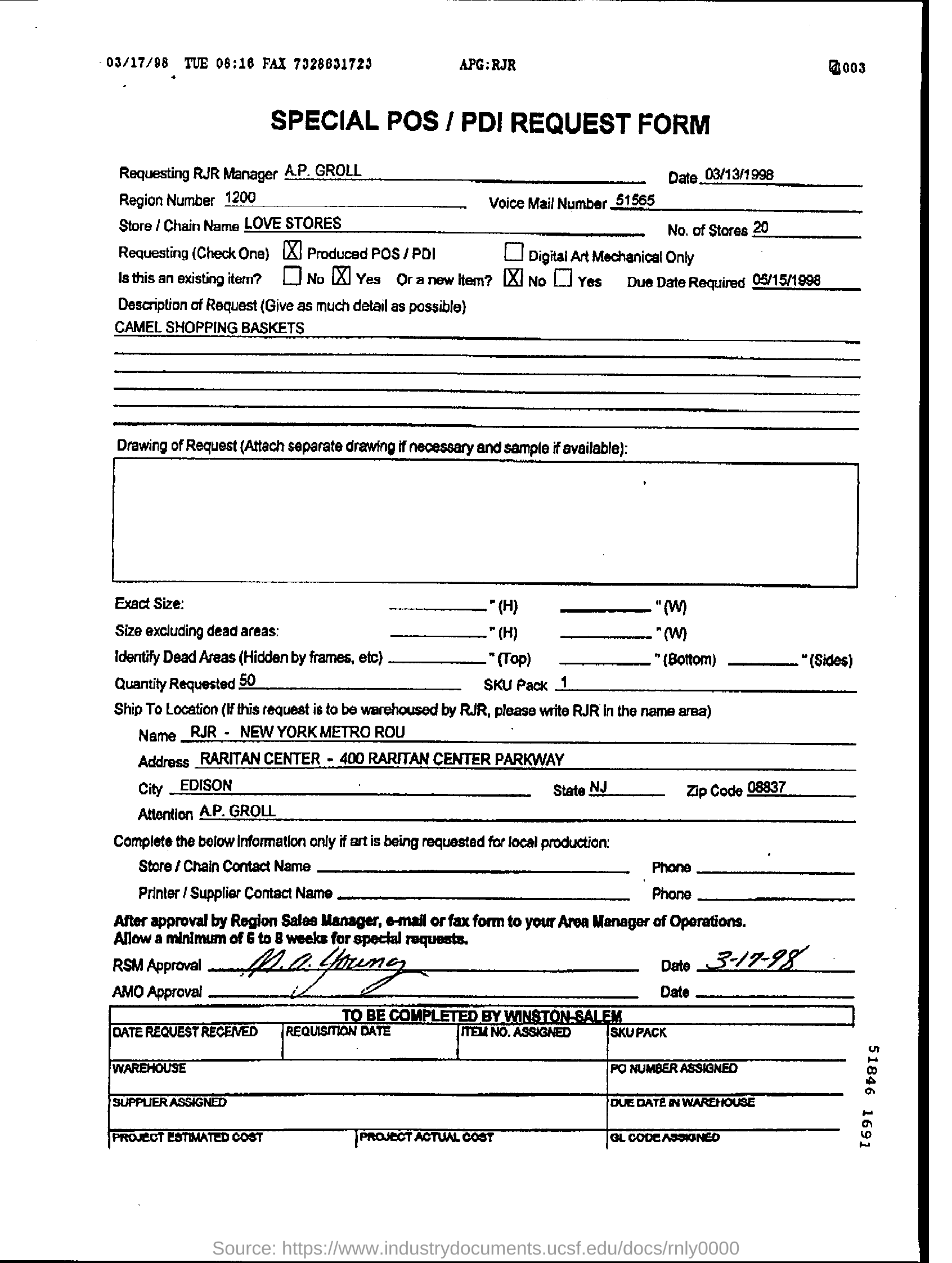} }
    
    \\
      \fbox{\includegraphics[width=0.8\linewidth,height=0.3\linewidth]{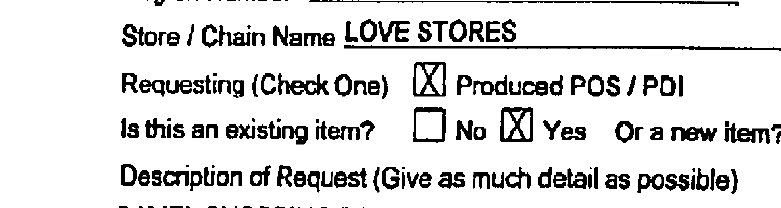}} 
      &
    \fbox{\includegraphics[width=0.8\linewidth,height=0.3\linewidth]{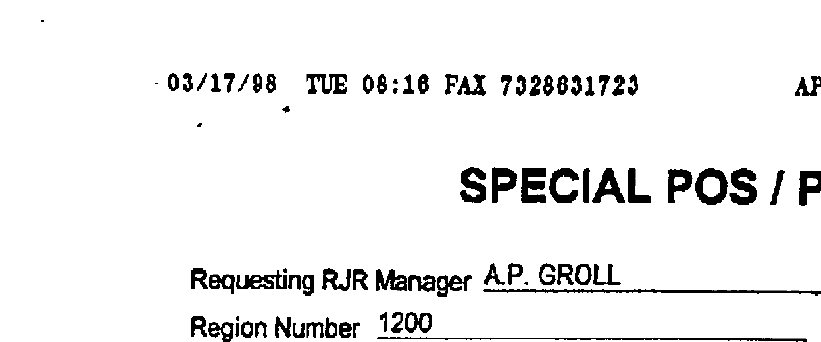}} 
     \\  \\
    \footnotesize{\fontfamily{qhv}\selectfont \textbf{Q: }Is it an existing item ?} \par 
    \footnotesize{\fontfamily{qhv}\selectfont \textbf{Question types: } \blue{\textit{form}} and \blue{\textit{yes/no}}} \par
    \footnotesize{\fontfamily{qhv}\selectfont \textbf{A: } yes}
    
    &
    
   \footnotesize{\fontfamily{qhv}\selectfont \textbf{Q: } What is the date given at the top left? } \par 
   \footnotesize{\fontfamily{qhv}\selectfont \textbf{Question types: } \blue{\textit{layout}} } \par

    \footnotesize{\fontfamily{qhv}\selectfont \textbf{A: }03/17/98}

    \\

\end{tabular}
\end{center}
\caption{On the left is a question based on an yes/no check box. On the right, the question seeks for  a date given at a particular spatial location --- top left of the page.}
\label{fig:question_type_examples yesno and layout}

\end{figure*}

\begin{figure*}[h]
\begin{center}
\begin{tabular}{p{0.4\linewidth}p{0.4\linewidth}}
 \includegraphics[width=1.3\linewidth]{images/rnly0000_2.png} 
    
    \\
      \fbox{\includegraphics[width=1.3\linewidth,height=0.3\linewidth]{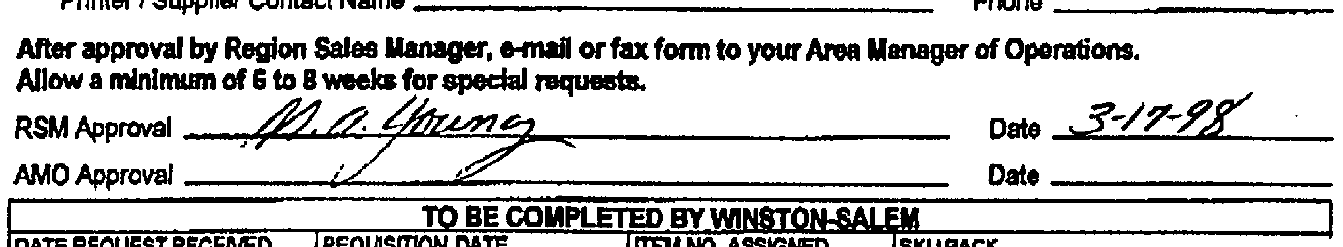}}

     \\  \\
    \footnotesize{\fontfamily{qhv}\selectfont \textbf{Q: }What is the date written next to RSM approval? } \par 
    \footnotesize{\fontfamily{qhv}\selectfont \textbf{Question types: } \blue{\textit{form}} and \blue{\textit{handwritten}}} \par
    \footnotesize{\fontfamily{qhv}\selectfont \textbf{A: }3-17-98}

    \\

\end{tabular}
\end{center}
\caption{Date is handwritten and it is shown in a  \textit{key:value} format.}
\label{fig:question_type_examples handwritten date form}

\end{figure*}

\begin{figure*}[h]
\begin{center}
\begin{tabular}{p{0.4\linewidth}p{0.4\linewidth}}
 \includegraphics[width=1.3\linewidth]{images/rnly0000_2.png} 
    
    \\
      \fbox{\includegraphics[width=1.3\linewidth,height=0.3\linewidth]{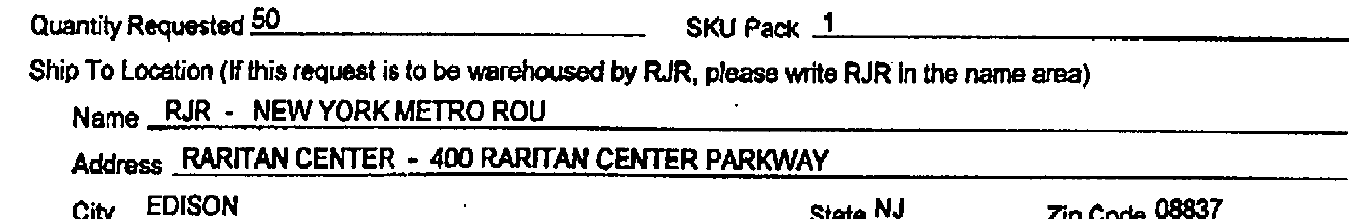}}

     \\  \\
    \footnotesize{\fontfamily{qhv}\selectfont \textbf{Q: }If the request needs to be warehoused by RJR, what needs to be done ?} \par 
    \footnotesize{\fontfamily{qhv}\selectfont \textbf{Question types: } \blue{\textit{running text}} } \par
    \footnotesize{\fontfamily{qhv}\selectfont \textbf{A: }write to RJR}

    \\

\end{tabular}
\end{center}
\caption{Question is grounded on a sentence.}
\label{fig:question_type running text}

\end{figure*}

\begin{figure*}[h]
\begin{center}
\begin{tabular}{p{0.4\linewidth}p{0.4\linewidth}}
 \multicolumn{2}{c}{\includegraphics[width=0.6\linewidth]{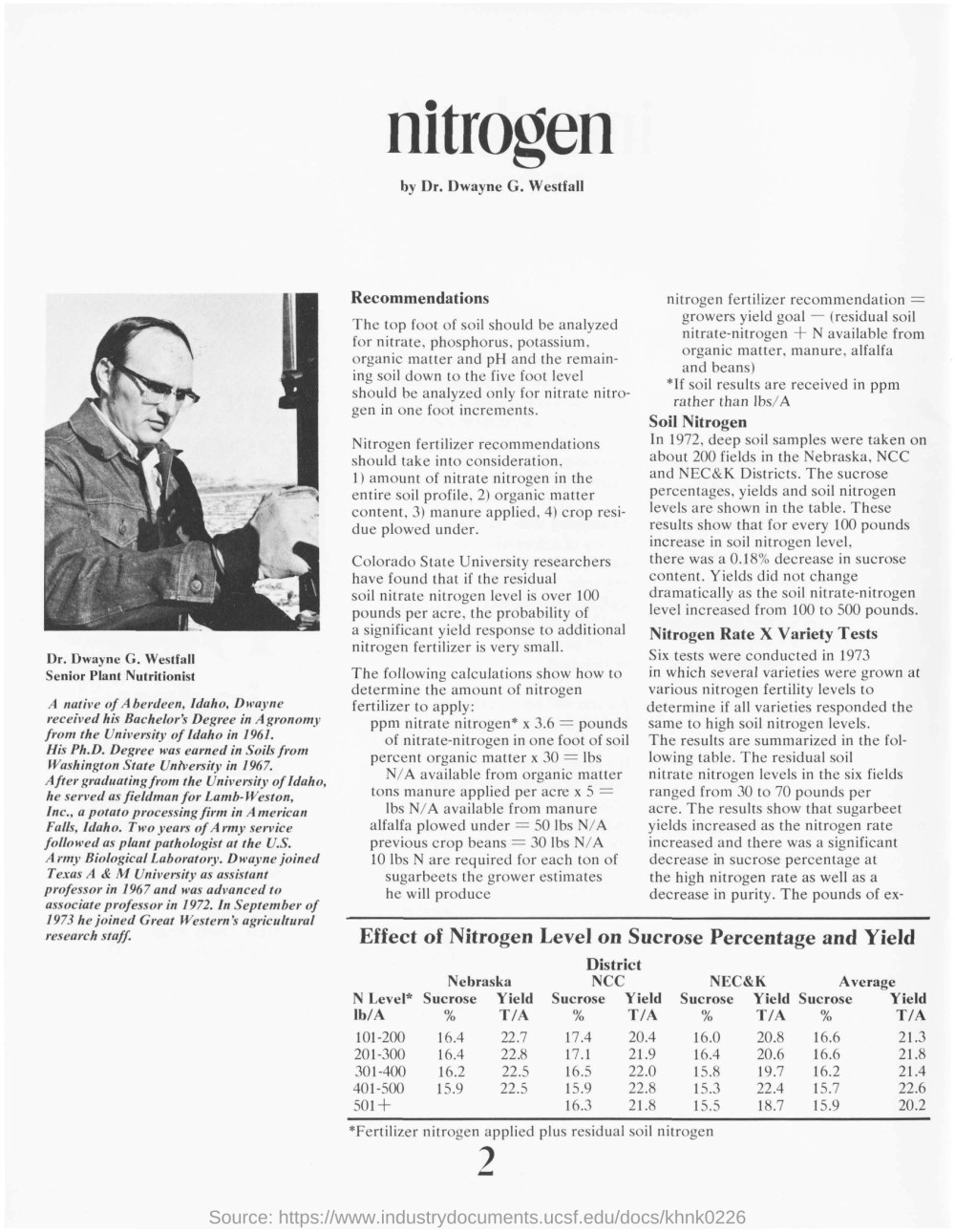} }
    
    \\
      \fbox{\includegraphics[width=0.8\linewidth,height=0.5\linewidth]{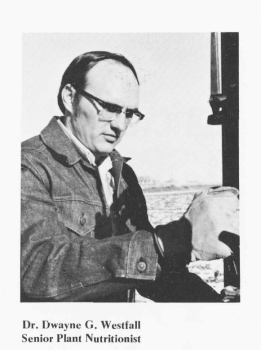}} 
      &
    \fbox{\includegraphics[width=1.1\linewidth,height=0.5\linewidth]{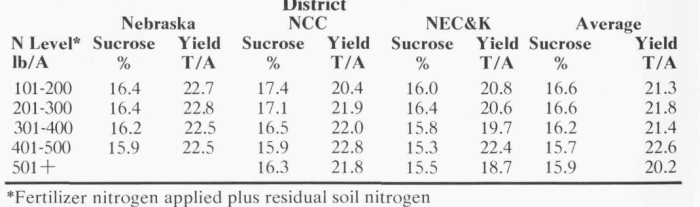}} 
     \\  \\
    \footnotesize{\fontfamily{qhv}\selectfont \textbf{Q:} Whose picture is given? } \par 
    \footnotesize{\fontfamily{qhv}\selectfont \textbf{Question types:} \blue{\textit{photograph}} and \blue{\textit{layout}}} \par
    \footnotesize{\fontfamily{qhv}\selectfont \textbf{A:} Dr. Dwayne G. Westfall}
    
    &
    
   \footnotesize{\fontfamily{qhv}\selectfont \textbf{Q: }What is the average sucrose \% for N level 501+ ?} \par 
   \footnotesize{\fontfamily{qhv}\selectfont \textbf{Question types: } \blue{\textit{table}} } \par

    \footnotesize{\fontfamily{qhv}\selectfont \textbf{A: }15.9}

    \\

\end{tabular}
\end{center}
\caption{On the left is a question asking for name of the person in the photograph. To answer the question on the right, one needs to parse the table and pick the value in the appropriate cell}
\label{fig:question_types photo and table}

\end{figure*}

\begin{figure*}[h]
\begin{center}
\begin{tabular}{p{0.4\linewidth}p{0.4\linewidth}}
 \includegraphics[width=1.3\linewidth]{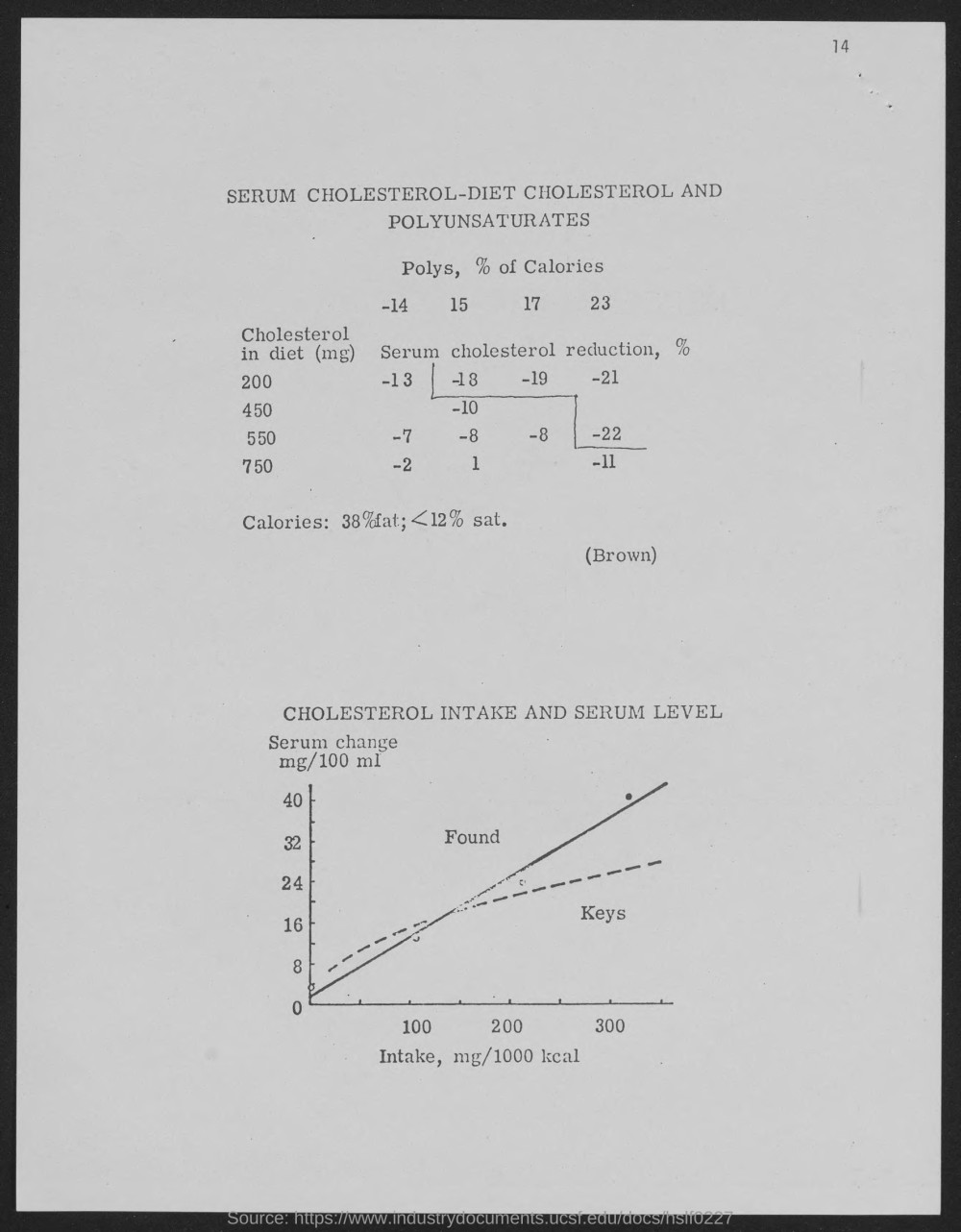} 
    
    \\
      \fbox{\includegraphics[width=1.3\linewidth,height=0.3\linewidth]{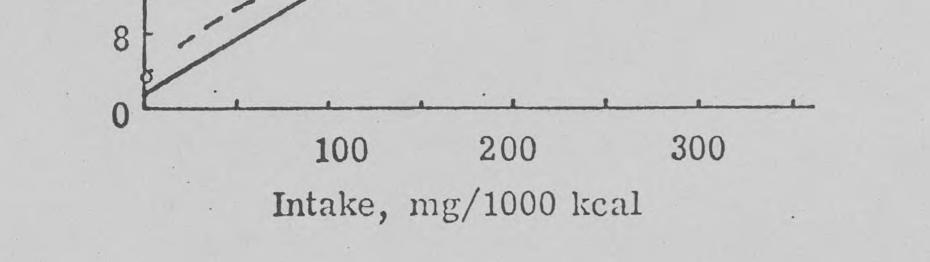}}

     \\  \\
    \footnotesize{\fontfamily{qhv}\selectfont \textbf{Q: } What is the highest value for ``Intake, mg/1000kcal" plotted on the `X' axis of the graph?} \par 
    \footnotesize{\fontfamily{qhv}\selectfont \textbf{Question types: } \blue{\textit{figure}} } \par
    \footnotesize{\fontfamily{qhv}\selectfont \textbf{A: }300}

    \\

\end{tabular}
\end{center}
\caption{Question is based on the plot shown at the bottom of the given image, asking for the highest value on the X axis}
\label{fig:question_type_examples figure}

\end{figure*}

Examples for types \textit{form}, \textit{yes/no} and \textit{layout} are shown in~\autoref{fig:question_type_examples yesno and layout}. Examples for a question based on a handwritten date in a form (types \textit{form} and \textit{handwritten}) are shown in~\autoref{fig:question_type_examples handwritten date form}. An example for a question based on information in the form of sentences or paragraphs ( type \textit{running text}) is shown in~\autoref{fig:question_type running text}. Examples for types \textit{photograph} and \textit{table} are shown in~\autoref{fig:question_types photo and table}. An example for a question based on a plot (type \textit{figure}) is shown in~\autoref{fig:question_type_examples figure}. In all examples a crop of the original image is shown below the original image, for better viewing of the image region where the question is based on.

\section{Additional Qualitative Examples}
\label{appendix:Additional Qualitative Examples}
Here we show more qualitative results from our baseline experiments. These results supplement the Results section (Section 5.3 ) in the main paper.

\begin{figure*}[h]
\begin{center}
\begin{tabular}{p{0.48\linewidth}p{0.48\linewidth}}
    \includegraphics[width=\linewidth]{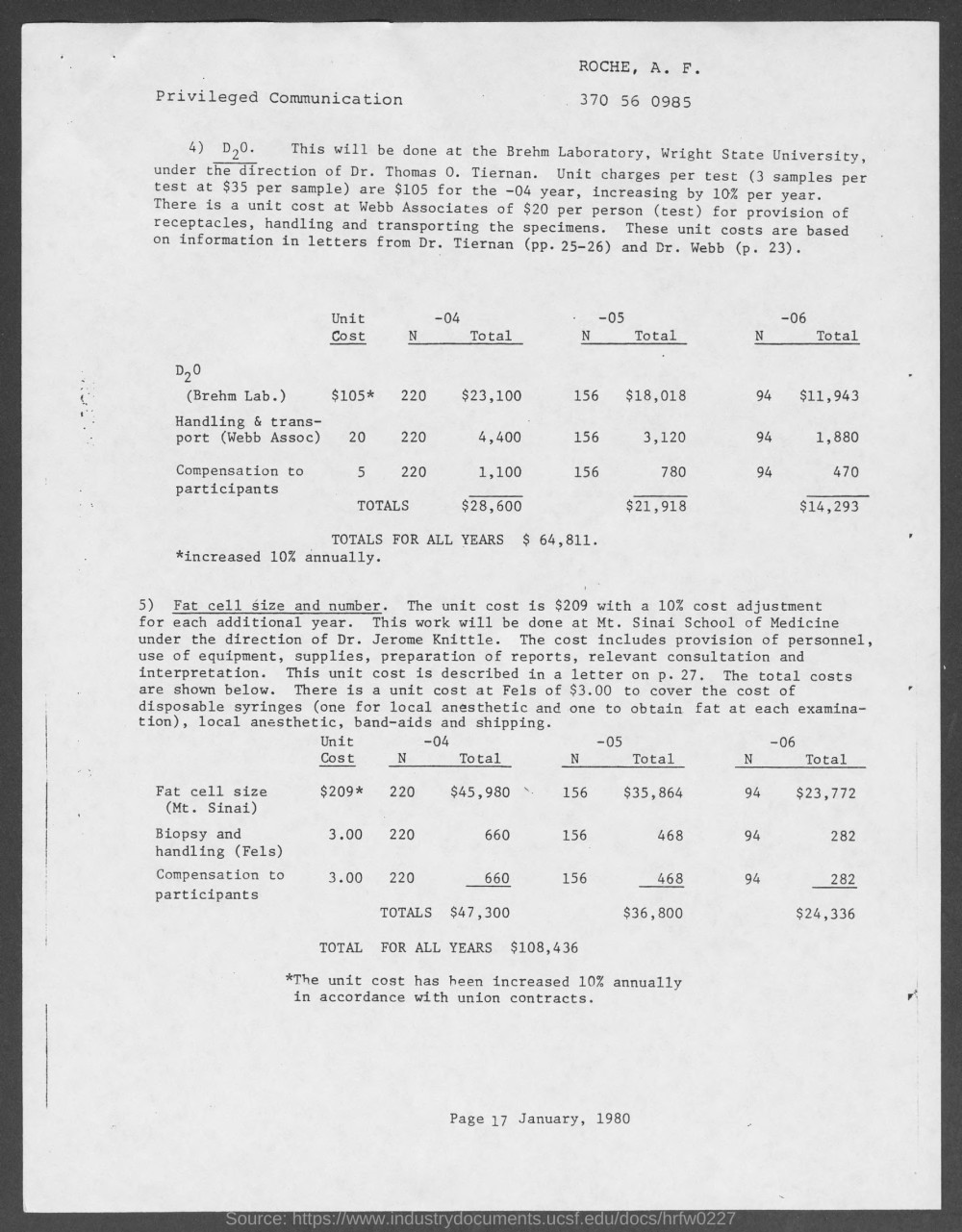} &
    \includegraphics[width=\linewidth]{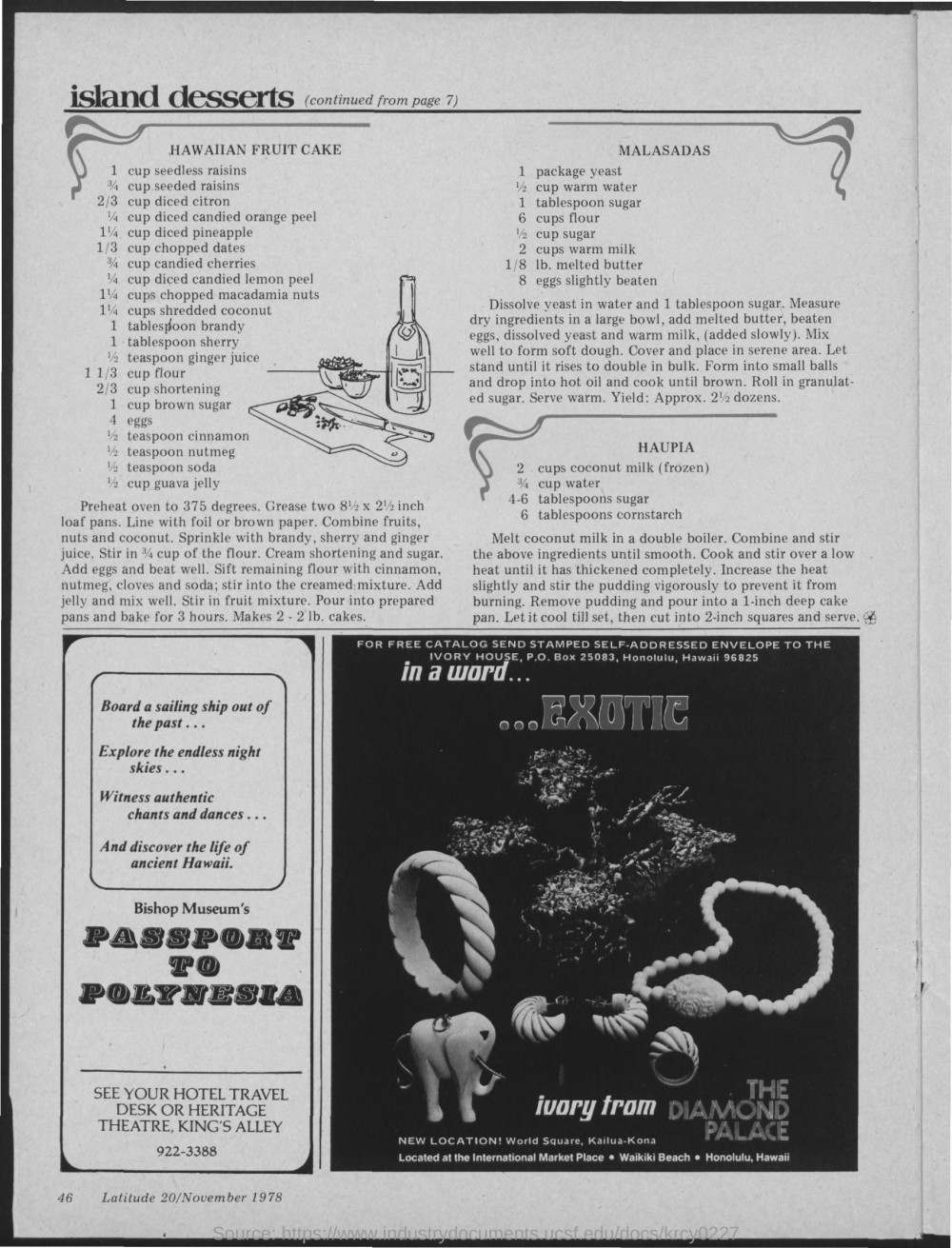}

    \\

  \fbox{\includegraphics[width=\linewidth,height=0.3\linewidth]{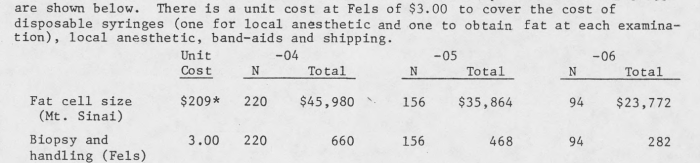}} &
  \fbox{\includegraphics[width=\linewidth,height=0.3\linewidth]{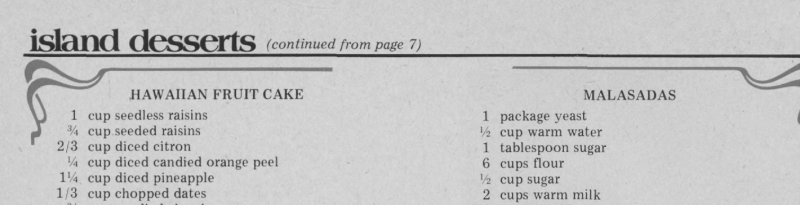}}

    \\
    \footnotesize{\fontfamily{qhv}\selectfont \textbf{Q: }What is the total cost for Fat cell size (Mt. SInai) in the -05 year ? } \par 
    \footnotesize{\fontfamily{qhv}\selectfont \textbf{GT: } \color{blue}\$35,864} 
    \par \footnotesize{\fontfamily{qhv}\selectfont \textbf{M4C best: } \color{red}4400} 
     \par \footnotesize{\fontfamily{qhv}\selectfont \textbf{BERT best: } \color{orange}\$35 , 864} 
     \par \footnotesize{\fontfamily{qhv}\selectfont \textbf{Human: } \color{green}\$35,864} 
    &

    \footnotesize{\fontfamily{qhv}\selectfont \textbf{Q: }What is the first recipe on the page? } \par  \footnotesize{\fontfamily{qhv}\selectfont \textbf{GT: } \color{blue}hawaiian fruit cake} 
    \par \footnotesize{\fontfamily{qhv}\selectfont \textbf{M4C best: } \color{red}island desserts (continued from cake} 
     \par \footnotesize{\fontfamily{qhv}\selectfont \textbf{BERT best: } \color{green}hawaiian fruit cake} 
     \par \footnotesize{\fontfamily{qhv}\selectfont \textbf{Human: } \color{green}hawaiian fruit cake}

\end{tabular}
\end{center}
\caption{\textbf{Examples where BERT QA model~\cite{bert} answers questions other than `running text' type.} On the left is a question based on a table and for the other question one needs to know the `first recipe' out of the two recipes shown. For the first question the model gets the answer correct except for an extra space, and in case of the second one the predicted answer matches exactly with the ground truth answer.  }
\label{fig:qual : bert wines}

\end{figure*}

\begin{figure*}[h]
\begin{center}
\begin{tabular}{p{0.48\linewidth}p{0.48\linewidth}}
    \includegraphics[width=\linewidth]{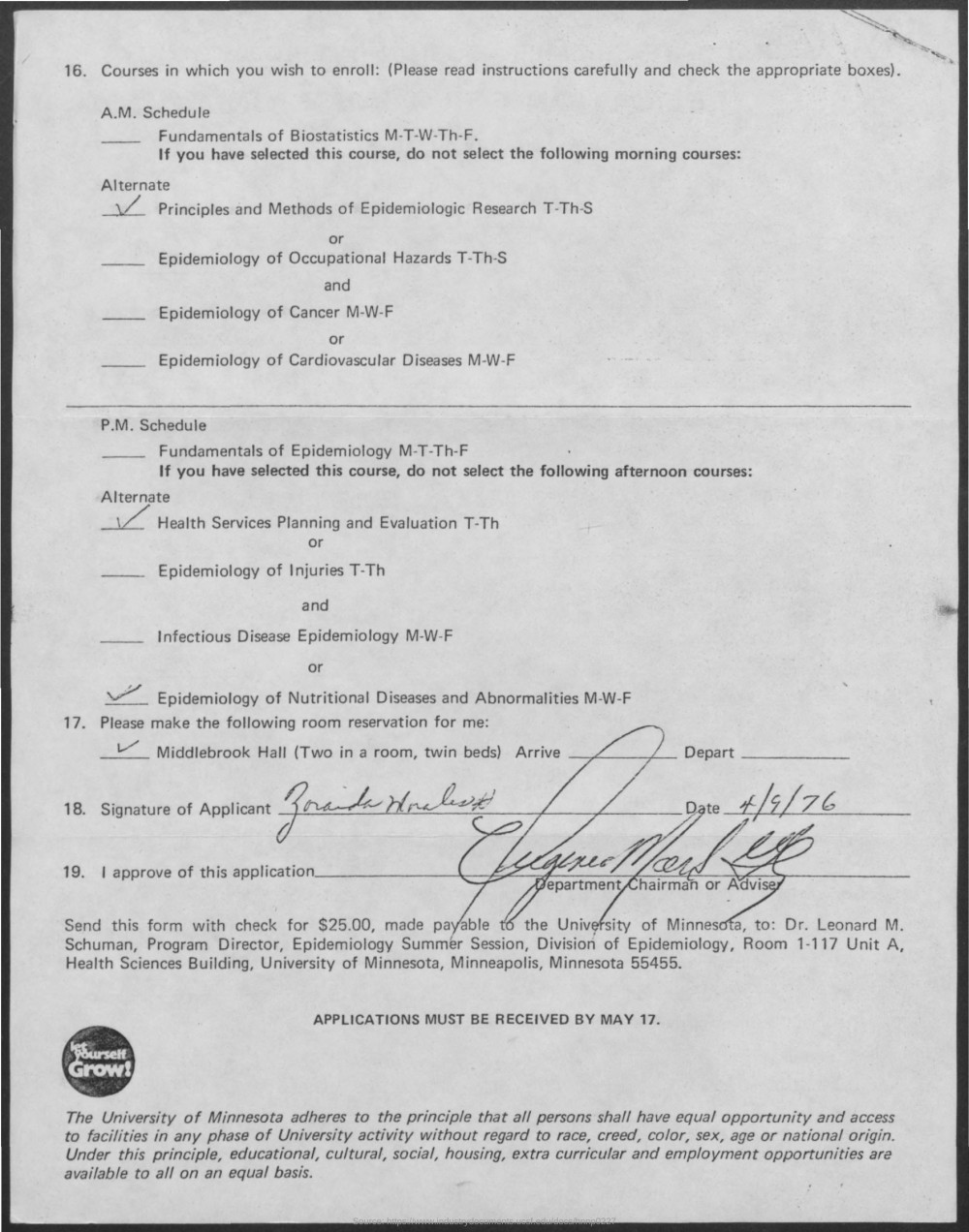} &
    \includegraphics[width=\linewidth]{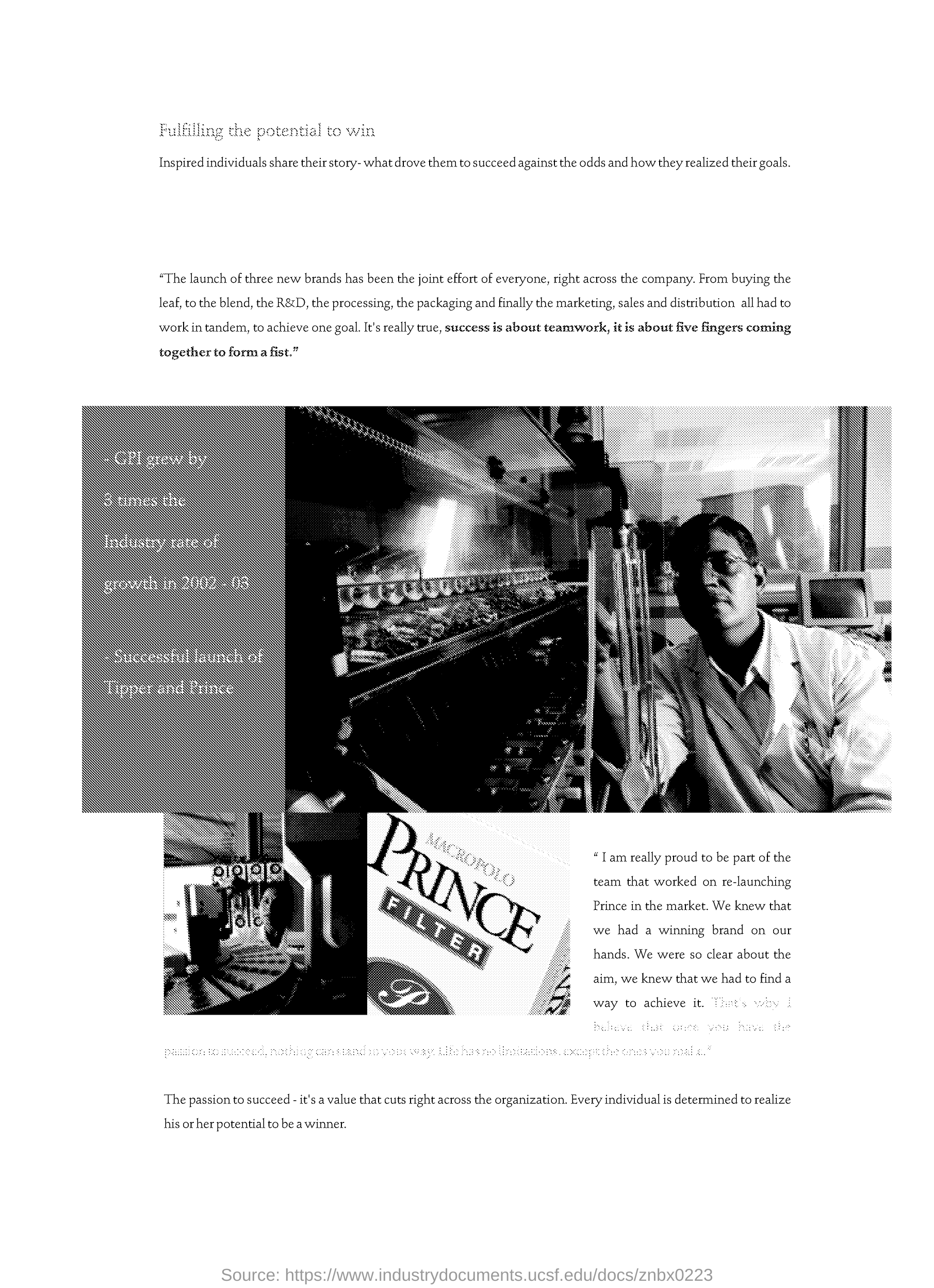}

    \\

  \fbox{\includegraphics[width=\linewidth,height=0.3\linewidth]{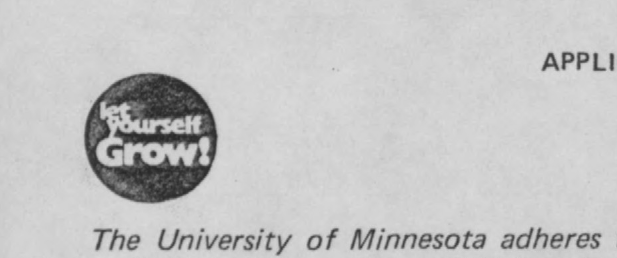}} &
  \fbox{\includegraphics[width=\linewidth,height=0.3\linewidth]{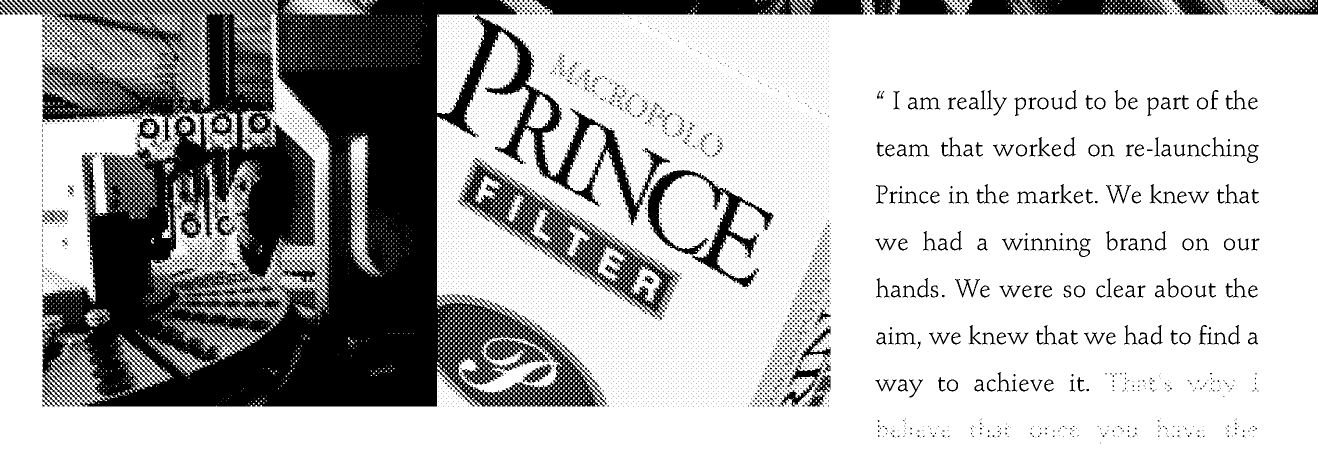}}

    \\

    \footnotesize{\fontfamily{qhv}\selectfont \textbf{Q: }What is written inside logo in the bottom of the document? } \par  \footnotesize{\fontfamily{qhv}\selectfont \textbf{GT: } \color{blue}let yourself grow!} 
    \par \footnotesize{\fontfamily{qhv}\selectfont \textbf{M4C best: } \color{orange} yourself grow!} 
     \par \footnotesize{\fontfamily{qhv}\selectfont \textbf{BERT best: } \color{red}$<no \quad  prediction>$} 
     \par \footnotesize{\fontfamily{qhv}\selectfont \textbf{Human: } \color{green}let yourself grow!} 
    &
    \footnotesize{\fontfamily{qhv}\selectfont \textbf{Q: }What Tobacco brand of GPI is shown in the picture? } \par \footnotesize{\fontfamily{qhv}\selectfont \textbf{GT: }\color{blue}Prince} 
    \par \footnotesize{\fontfamily{qhv}\selectfont \textbf{M4C best: } \color{green}prince} 
     \par \footnotesize{\fontfamily{qhv}\selectfont \textbf{BERT best: } \color{red}$<no\quad prediction>$} 
     \par \footnotesize{\fontfamily{qhv}\selectfont \textbf{Human: } \color{green}prince}

\end{tabular}
\end{center}
\caption{\textbf{How does the \mc~\cite{m4c} model perform on questions based on pictures or photographs.}
Here we show two examples where the best variant of the \mc model outperform the BERT best model in answering `layout' type questions seeking to read what is written in a logo/pack. The BERT model doesnt make any predictions for the questions.}
\label{fig:qual : m4c_wins}

\end{figure*}

\begin{figure*}[h]
\begin{center}
\begin{tabular}{p{0.4\linewidth}p{0.4\linewidth}}
    \multicolumn{2}{c}{\includegraphics[width=0.6\linewidth]{images/tnbx0223_15.png}}

    \\

   \multicolumn{2}{c}{\fbox{\includegraphics[width=0.3\linewidth,height=0.2\linewidth]{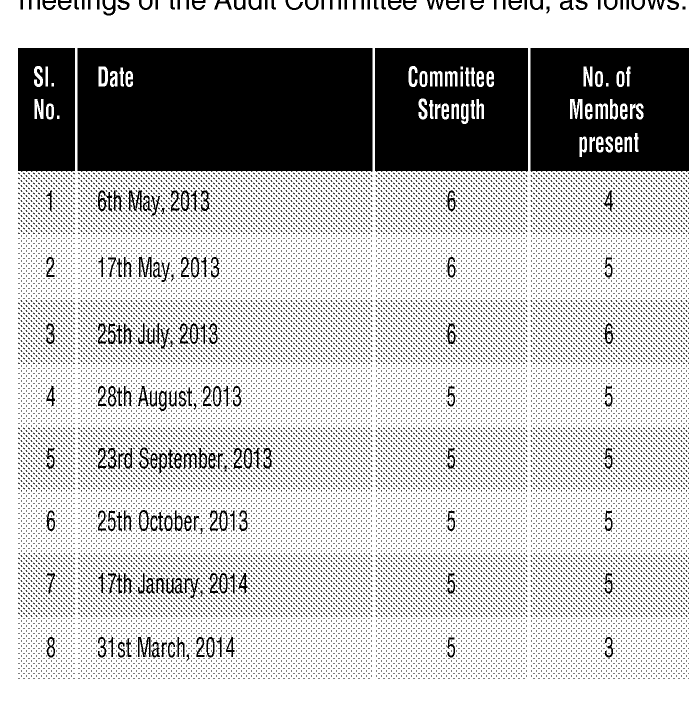}}}

    \\
    \\
    \footnotesize{\fontfamily{qhv}\selectfont \textbf{Q: }What was the committee strength for the first meeting? } \par \footnotesize{\fontfamily{qhv}\selectfont \textbf{GT: } \color{blue}6} 
    \par \footnotesize{\fontfamily{qhv}\selectfont \textbf{M4C best: } \color{green}6} 
     \par \footnotesize{\fontfamily{qhv}\selectfont \textbf{BERT best: } \color{green}6} 
     \par \footnotesize{\fontfamily{qhv}\selectfont \textbf{Human: } \color{green}6} 
    &

    \footnotesize{\fontfamily{qhv}\selectfont \textbf{Q: }What was the committee strength for the last meeting? } \par  \footnotesize{\fontfamily{qhv}\selectfont \textbf{GT: } \color{blue}5} 
    \par \footnotesize{\fontfamily{qhv}\selectfont \textbf{M4C best: } \color{red}6} 
     \par \footnotesize{\fontfamily{qhv}\selectfont \textbf{BERT best: } \color{red}6} 
     \par \footnotesize{\fontfamily{qhv}\selectfont \textbf{Human: } \color{green}5}

\end{tabular}
\end{center}
\caption{\textbf{Contrasting results for similar questions.} Here both the questions are based on the table at the bottom of the image. Both questions ask for `committee strength' for a particular meeting (first or last). Both models get the answer right for the first one. But for the question on the right, the models predict same answer as the first one (``6'') while the  ground truth is ``5''. This suggests that the models' predictions are not backed by a proper reasoning/grounding in all cases.}
\label{fig:qual : inconsistent}

\end{figure*}

\begin{figure*}[h]
\begin{center}
\begin{tabular}{p{0.48\linewidth}p{0.48\linewidth}}
    \includegraphics[width=\linewidth]{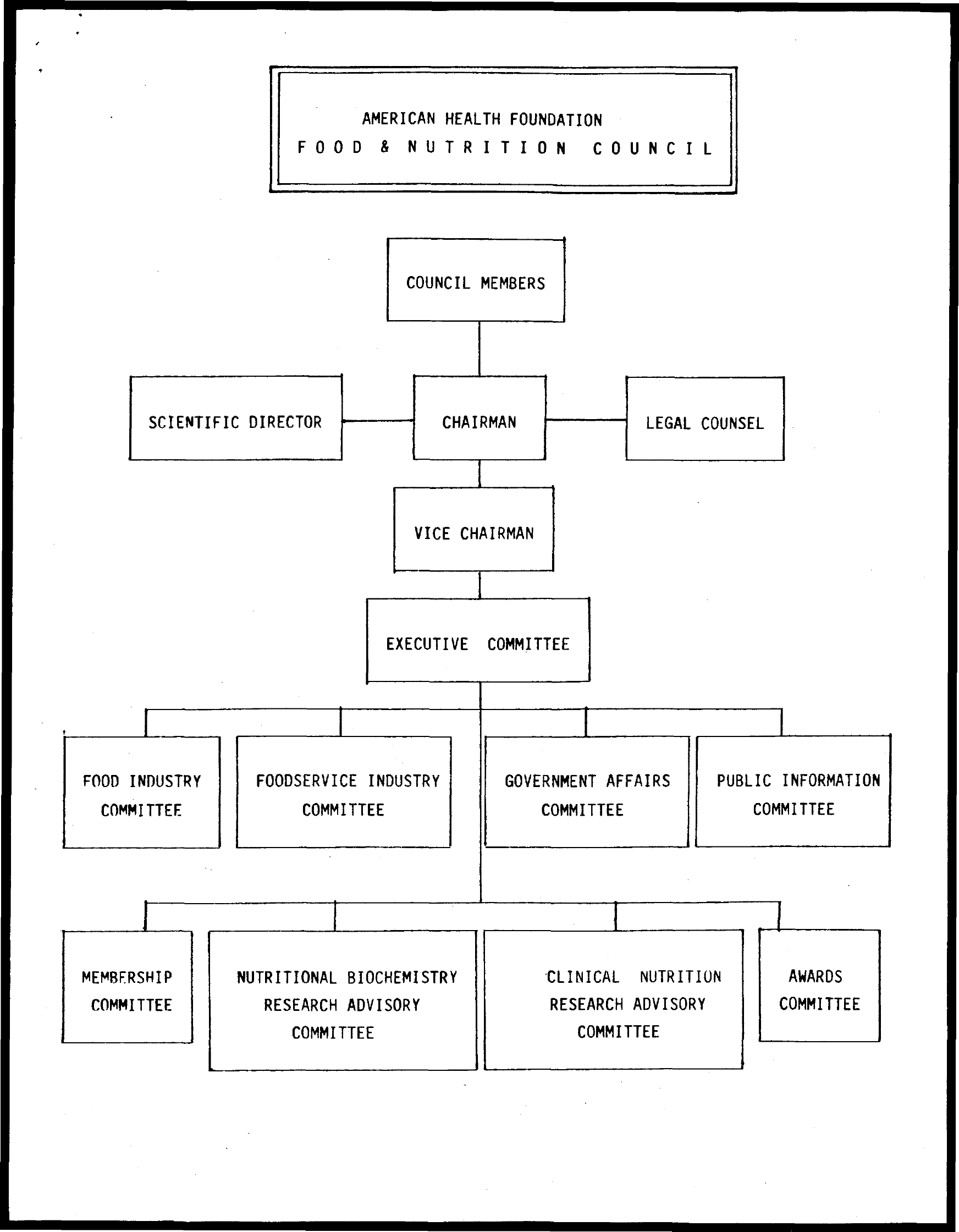} &
    \includegraphics[width=\linewidth]{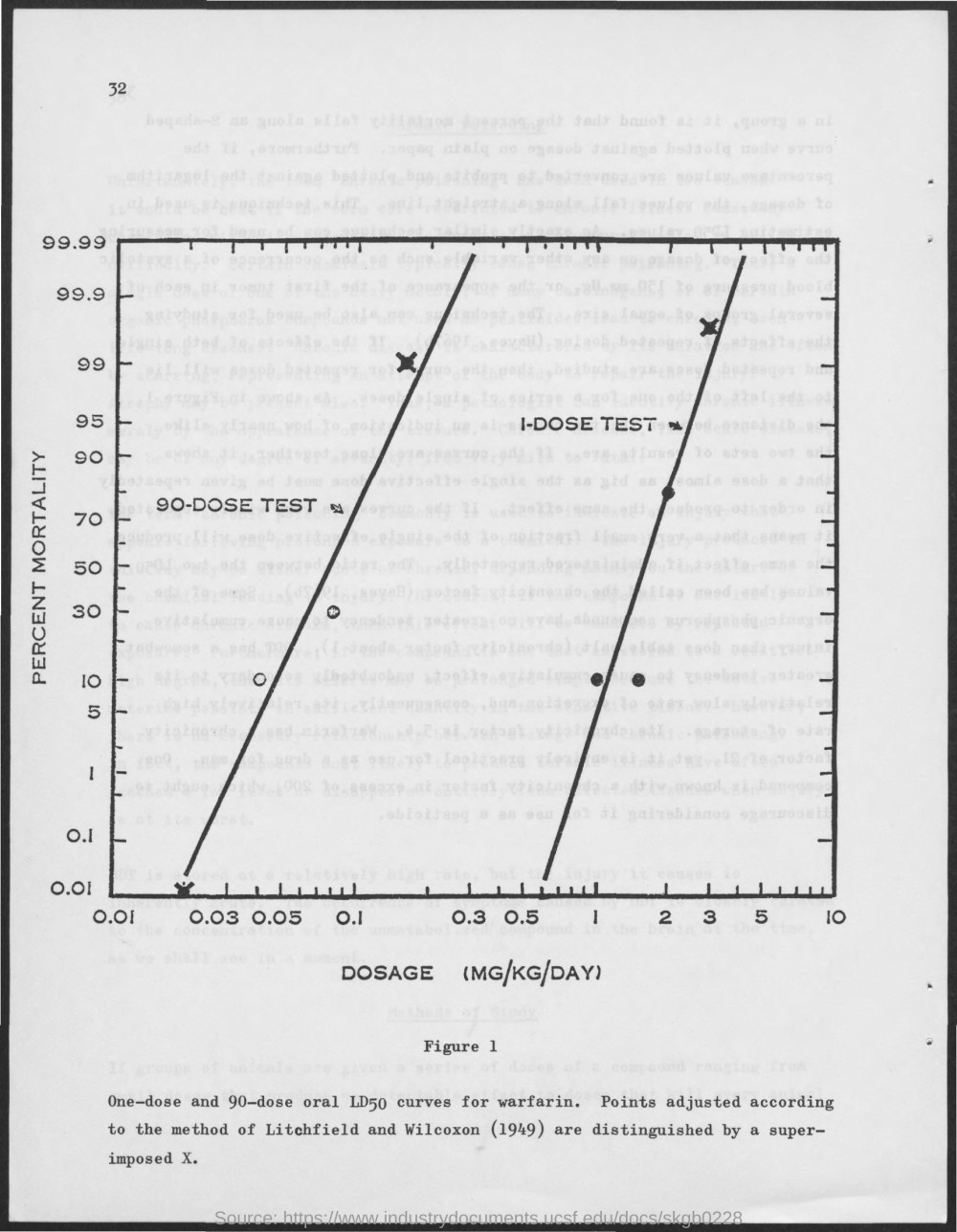}

    \\

  \fbox{\includegraphics[width=\linewidth,height=0.3\linewidth]{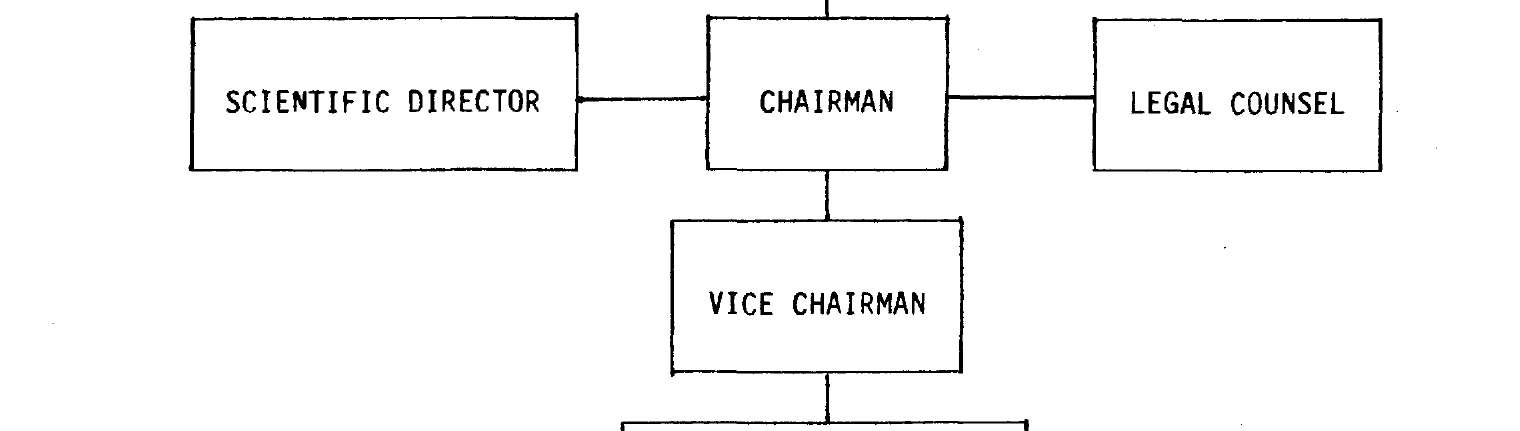}} &
  \fbox{\includegraphics[width=\linewidth,height=0.3\linewidth]{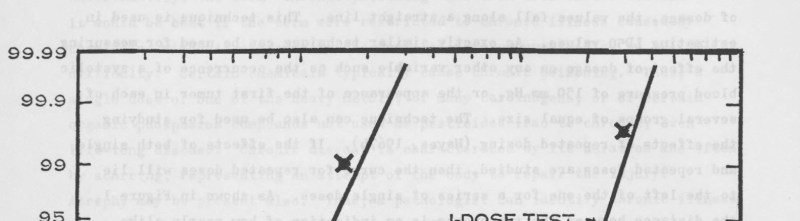}}

   \\
    \footnotesize{\fontfamily{qhv}\selectfont \textbf{Q: }What is the position above "vice chairman" ? } \par 
    \footnotesize{\fontfamily{qhv}\selectfont \textbf{GT: } \color{blue}chairman} 
    \par \footnotesize{\fontfamily{qhv}\selectfont \textbf{M4C best: } \color{red}legal counsel} 
     \par \footnotesize{\fontfamily{qhv}\selectfont \textbf{BERT best: } \color{red}legal counsel} 
     \par \footnotesize{\fontfamily{qhv}\selectfont \textbf{Human: } \color{green}chairman} 
    &

    \footnotesize{\fontfamily{qhv}\selectfont \textbf{Q: }What is the highest value shown on the vertical axis? } \par  \footnotesize{\fontfamily{qhv}\selectfont \textbf{GT:} \color{blue}99.99} 
    \par \footnotesize{\fontfamily{qhv}\selectfont \textbf{M4C best: } \color{red}50} 
     \par \footnotesize{\fontfamily{qhv}\selectfont \textbf{BERT best: } \color{red}32} 
     \par \footnotesize{\fontfamily{qhv}\selectfont \textbf{Human: } \color{green}99.99}

\end{tabular}
\end{center}
\caption{\textbf{Understanding figures and diagrams}. In case of the question on the left, one needs to understand an organizational hierarchy diagram. For the second question, one needs to know what a `vertical axis' is, and then find the largest value. Both the  models fail to answer the questions.}
\label{fig: qual: reasoning}

\end{figure*}

\begin{figure*}[h]
\begin{center}
\begin{tabular}{p{0.48\linewidth}p{0.48\linewidth}}
    \includegraphics[width=\linewidth]{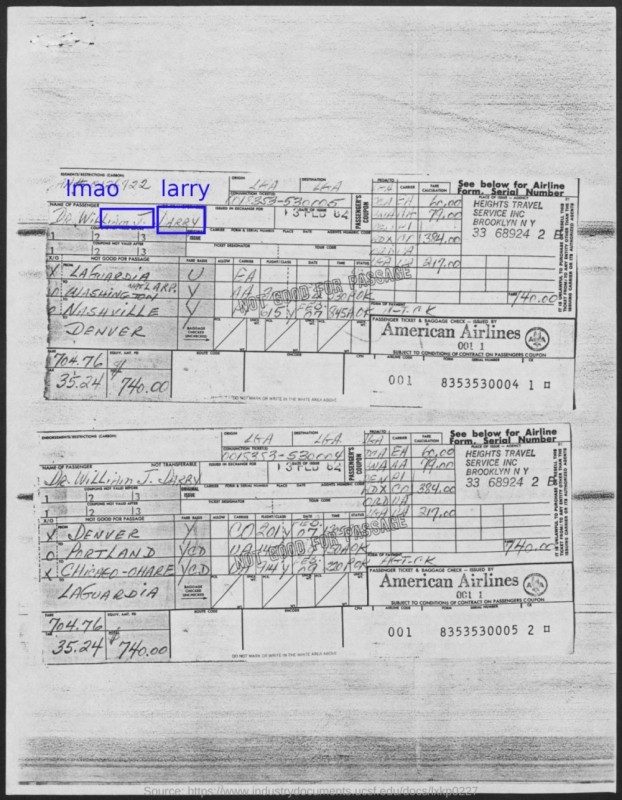} &
    \includegraphics[width=\linewidth]{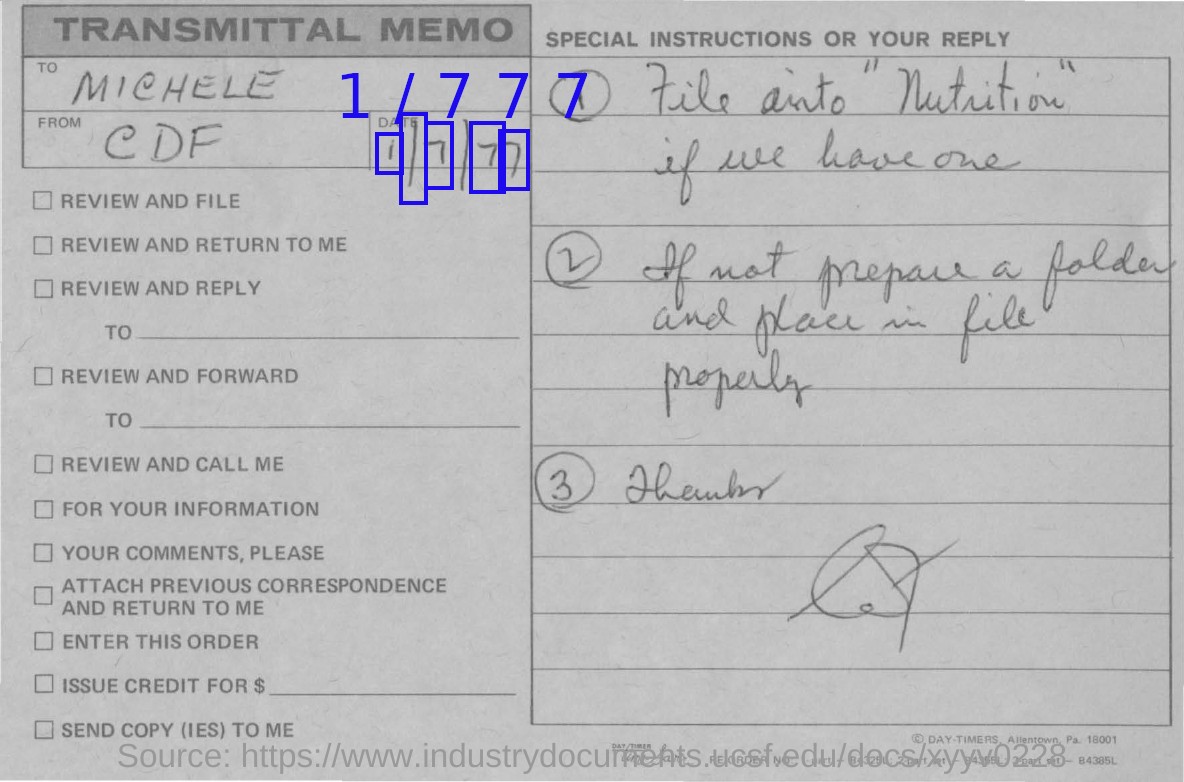}

    \\

  \fbox{\includegraphics[width=\linewidth,height=0.3\linewidth]{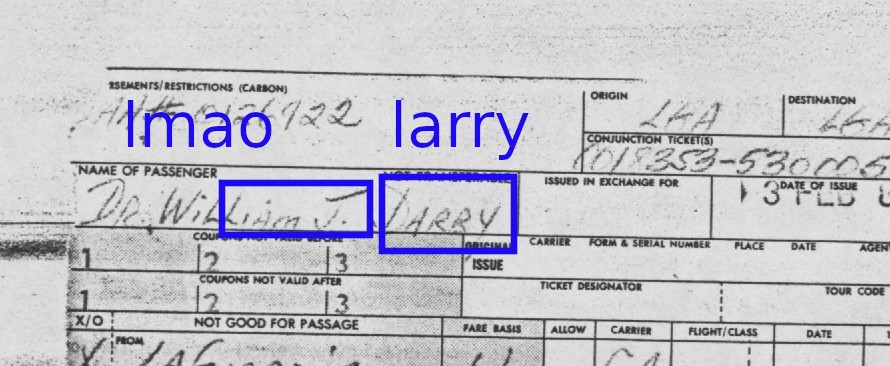}} &
  \fbox{\includegraphics[width=\linewidth,height=0.3\linewidth]{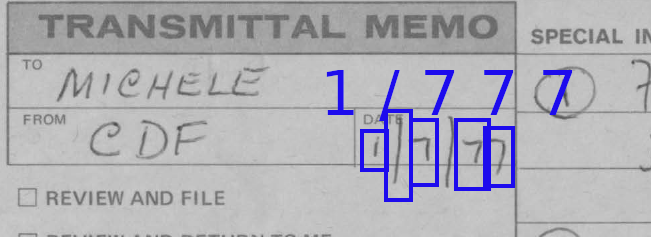}}

   \\
    \footnotesize{\fontfamily{qhv}\selectfont \textbf{Q: }What is the name of the passenger? } \par  \footnotesize{\fontfamily{qhv}\selectfont \textbf{GT: } \color{blue}dr. william j. darby} 
    \par \footnotesize{\fontfamily{qhv}\selectfont \textbf{M4C best: } \color{red}larry} 
     \par \footnotesize{\fontfamily{qhv}\selectfont \textbf{BERT best:} \color{red}larry} 
     \par \footnotesize{\fontfamily{qhv}\selectfont \textbf{Human: } \color{orange}dr. william j. darry} 
    &
   
    \footnotesize{\fontfamily{qhv}\selectfont \textbf{Q: }What is the date present in the memo ?} \par 
    \footnotesize{\fontfamily{qhv}\selectfont \textbf{GT:} \color{blue}1/7/77} 
    \par \footnotesize{\fontfamily{qhv}\selectfont \textbf{M4C best: } \color{red}1 7 77} 
     \par \footnotesize{\fontfamily{qhv}\selectfont \textbf{BERT best: } \color{red}1 / 7} 
     \par \footnotesize{\fontfamily{qhv}\selectfont \textbf{Human: } \color{green}1/7/77}

\end{tabular}
\end{center}
\caption{\textbf{Impact of OCR errors.} Here the models are able to ground the questions correctly on the relevant information in the image, but failed to get the answers correct owing to the OCR errors. In case of the question on the left, even the answer entered by the human volunteer is not exactly matching with the ground truth. In case of the second question, OCR has split the date into multiple tokens due to over segmentation, resulting in incorrect answers by both the models. }
\label{fig: qual: ocr error}

\end{figure*}

Remember that BERT~\cite{bert} question answering model is designed to answer questions asked on sentences or paragraphs of text ( reading comprehension). In~\autoref{fig:qual : bert wines} we show two examples where the model answers questions outside the ambit of reading comprehension style question answering. In~\autoref{fig:qual : m4c_wins} we show examples where the \mc~\cite{m4c} model outperforms the BERT model to answer questions based on text seen on pictures or photographs. Such questions are  similar to questions in \textvqa~\cite{textvqa} or \stvqa~\cite{stvqa_iccv} datasets where \mc model yield state-of-the-art results. In~\autoref{fig:qual : inconsistent} we show an example where both the models yield inconsistent results when posed with questions of similar nature, highlighting lack of reasoning behind answering. In~\autoref{fig: qual: reasoning} we show two examples where both the \mc and BERT model fail to answer questions which require understanding of a  figure or a diagram. In~\autoref{fig: qual: ocr error} we show how OCR errors have resulted in wrong answers although the models manage to ground the questions correctly.


\end{document}